\def\hycoarxiv{}
\documentclass[11pt]{article}
\ifdefined\hycoarxiv
  \usepackage[preprint]{acl}
\else
  \usepackage[review]{acl}
\fi

\usepackage{times}
\usepackage{latexsym}
\usepackage[T1]{fontenc}
\usepackage[utf8]{inputenc}
\usepackage{microtype}
\usepackage{inconsolata}
\usepackage{graphicx}
\usepackage{booktabs}
\usepackage{array}
\newcolumntype{L}[1]{>{\raggedright\arraybackslash}p{#1}}
\usepackage{multirow}
\usepackage{amsmath}
\usepackage{amssymb}
\usepackage{xcolor}
\usepackage{xspace}
\usepackage{pifont}
\usepackage{algorithm}
\usepackage[noend]{algpseudocode}
\newcommand{\cmark}{\ding{51}}
\newcommand{\nomark}{\textcolor{black!35}{--}}

\newif\ifdraftmarks
\draftmarksfalse
\ifdefined\hycoarxiv \draftmarksfalse \fi
\ifdraftmarks
  \newcommand{\todo}[1]{\textcolor{red}{[TODO: #1]}}
  \newcommand{\evid}[1]{\textcolor{blue}{[#1]}}
  \newcommand{\tbd}[1][]{\textcolor{red}{\textbf{?}#1}}
\else
  \newcommand{\todo}[1]{\unskip}
  \newcommand{\evid}[1]{\unskip}
  \newcommand{\tbd}[1][]{\unskip}
\fi
\newif\ifwithappendix
\withappendixtrue
\ifdefined\hycoarxivnoappendix \withappendixfalse \fi
\ifwithappendix
  \DeclareRobustCommand{\appx}[1]{#1}
\else
  \DeclareRobustCommand{\appx}[1]{\unskip}
\fi
\newcommand{\sysname}{\textsc{HyCo}\xspace}
\newcommand{\sysnametwo}{\mbox{\textsc{HyCo}-2P}\xspace}
\newcommand{\verifier}{verifier\xspace}

\newcommand{\slmdrafter}{SLM Drafter\xspace}

\newcommand{\mtpdrafter}{MTP Drafter\xspace}

\newcommand{\hybriddrafter}{Hybrid Drafter\xspace}

\newcommand{\serverhead}{server head\xspace}

\newcommand{\infsum}{$\infty$Bench Sum\xspace}
\newcommand{\epsacc}{$\epsilon$-threshold acceptance\xspace}
\xspaceaddexceptions{]}
\ifdraftmarks
  
\else
  
\fi

\newcommand{\hyTestSaveSB}{38.6}
\newcommand{\hyTestSaveHm}{29.2}
\newcommand{\hyTestTokSB}{35.0}

\newcommand{\hyTestHeadLoss}{4.0}
\newcommand{\hyTestHeadLossLo}{3.0}
\newcommand{\hyTestHeadLossHi}{4.9}

\newcommand{\hyOnDeviceTok}{26.4}

\newcommand{\hyEps}{0.005}
\newcommand{\hyDevTokHyco}{12.1}
\newcommand{\hyDevTokSlm}{36.8}

\title{\textsc{HyCo}: Hybrid Drafting for Cost-Efficient\\Mobile--Cloud Speculative Decoding}

\ifdefined\hycoarxiv
  \author{Hanbin Bae \quad Heejin Choi \quad Joohyung Lee \\
    \textbf{Byoung Jin Choi \quad Sangjun Park \quad Hoon-Young Cho} \\
    AI Solution Team, Samsung Research, Seoul, Republic of Korea \\
    {\normalsize\texttt{\{bhb0722.bae,hee\_jin.choi,jooooh.lee,bj87.choi,sj0.park,h.y.cho\}@samsung.com}}}
\else
  \author{Anonymous ACL submission}
\fi

\begin{document}
\maketitle

\begin{abstract}
Device--cloud speculative decoding reduces server computation, but growing a deep draft tree with a small language model (SLM) requires repeated passes on a mobile neural processing unit (NPU).
We present \sysname, which runs the server model's native multi-token prediction (MTP) head on the device to build a deep chain and uses one SLM pass to add alternative branches.
The server returns the hidden state, input embedding, and key--value entries needed by the head instead of executing its sequential steps.
Quantization, a restricted vocabulary, and a 1K-token window fit both drafters to the NPU.
An $\epsilon$-threshold rule additionally relaxes exact preservation of the verifier's sampling distribution.
In single-client runs with a Galaxy S26 Ultra over LTE/5G, \sysname\ reduces server GPU time per output token relative to server-side MTP self-speculation by \hyTestSaveSB\% on Spec-Bench and \hyTestSaveHm\% on 32K-token summarization.
With the same acceptance rule, three runs of each system on Spec-Bench show 9.0 percentage points more saving and $1.9\times$ delivery speed than an SLM-only drafter.
\sysname\ differs from Cloud+MTP by $+0.2$ judge-score points (paired 95\% CI: $-0.9$ to $+1.3$); exploratory translation favors SpecEdge-$\epsilon$.
Without relaxation, \sysname\ exact saves 27.7\% and 11.9\% on the two datasets.
\end{abstract}

\section{Introduction}
\label{sec:intro}

Speculative decoding accelerates large language models (LLMs) by verifying several draft tokens in one forward pass; exact verification preserves the LLM's sampling distribution~\citep{leviathan2023fast,chen2023accelerating}.
Serving engines support this approach~\citep{kwon2023efficient,zheng2024sglang}, and recent LLMs include native multi-token prediction (MTP) heads for server-side self-speculation~\citep{deepseekai2024v3,gemmateam2026gemma4}.
\emph{Device--cloud speculative decoding} instead places a small language model (SLM) on the user's device and retains the verifier on the server~\citep{park2025specedge,li2025sled}.
This division can reduce server work, but communication can make an individual request slower than server-only speculation~\citep{lyu2026distance}.
The goal is therefore to save server computation while maintaining a practical mobile delivery speed.

We evaluate this trade-off against two distinct reference points.
On the server, the goal is to use less GPU time per output token than \emph{Cloud+MTP}, which already runs the LLM's native MTP self-speculation.
On the device, the goal is to deliver tokens at least as fast as the standalone SLM used in our evaluation.
Cloud+MTP therefore measures the remaining server-side opportunity, whereas the standalone SLM provides a local alternative for the user.
These goals concern server execution time and user-visible delivery, respectively, rather than a single shared speedup metric.

The main obstacle is the cost of obtaining useful draft-tree depth on a mobile neural processing unit (NPU).
A tree provides alternative tokens at each position and can be verified in one pass~\citep{miao2024specinfer}, but an autoregressive SLM requires another sequential pass for each additional level.
Higher acceptance can amortize server work and communication across more output tokens, although acceptance varies across tasks~\citep{liu2024turbospec,xia2024unlocking}.
In our deployment, the four-level SpecEdge-style tree increases the mean number of committed output tokens per round (MAT, including the bonus token), but its four SLM passes make delivery slower than standalone local generation.
Integer-graph, memory, and context constraints further limit the device.
The challenge is to obtain useful depth and alternatives without repeatedly executing the SLM.

\begin{figure}[t]
  \centering
  \includegraphics[width=\columnwidth]{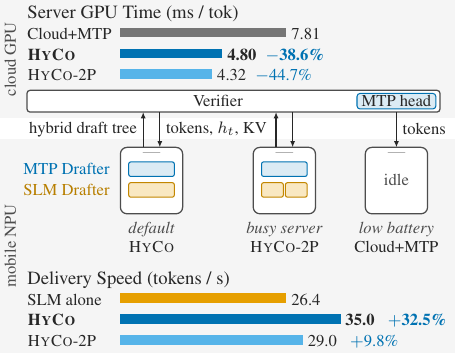}
  \caption{\sysname\ on Spec-Bench: server GPU time relative to exact Cloud+MTP and delivery rate relative to the standalone SLM. \sysname\ and \sysnametwo\ use $\epsilon{=}\hyEps$; \sysnametwo\ uses two SLM passes. Italic labels denote illustrative, not evaluated, deployment priorities. KV: key--value. \sysname\ shows a three-run mean; other results use one run.}
  \label{fig:overview}
\end{figure}

We present \sysname, which addresses this bottleneck by dividing drafting work between two models on the device (Figure~\ref{fig:overview}).
The verifier's native MTP head, called the \emph{\mtpdrafter}, supplies a sequential chain that gives the tree its depth.
In our deployment, the head reads only a small slice of verifier state, which the server returns after verification instead of executing the head's steps itself.
The \emph{\slmdrafter} then uses one pass to add alternatives under the root and the MTP-generated nodes, allowing the tree to continue when a chain token is not accepted.
Together, these models form the \emph{\hybriddrafter} and produce a hybrid draft tree.
The head remains sequential; the change is that increasing the chain's depth no longer requires additional SLM passes.

This division of drafting work supports exact verification.
Quantization, a restricted output vocabulary, and a 1K-token window make both drafters fit the NPU, but can reduce acceptance.
Under exact verification, these proposal restrictions do not change the verifier's sampling distribution.
Default \sysname\ additionally uses \emph{\epsacc}: it admits drafted tokens whose probability under the verifier's sampling distribution reaches $\epsilon$~\citep{hewitt2022truncation}.
This separate choice raises acceptance but is lossy.
We therefore evaluate the drafting design with exact verification as well as with the default relaxed rule, and compare it with an SLM-only tree under the same rule while assessing output quality separately.

We evaluate Gemma~4 26B-A4B on an RTX PRO 6000 cloud GPU, with its MTP head and Gemma~4 E2B on a Galaxy S26 Ultra over LTE/5G.
In single-client runs, \sysname\ exact reduces server GPU time per output token relative to exact Cloud+MTP by 27.7\% on Spec-Bench~\citep[SB;][]{xia2024unlocking} and 11.9\% on 32K-token summarization from \infsum~\citep{zhang2024infinitebench}.
Default \sysname, with \epsacc, increases these savings to \hyTestSaveSB\% and \hyTestSaveHm\%.
On SB, three runs under the same relaxed rule yield savings of $38.6\pm0.2$\% for \sysname\ and $29.6\pm0.2$\% for SpecEdge-$\epsilon$, the SLM-only baseline ($\pm$: run SD).
\sysname\ delivers \hyTestTokSB\ tokens/s, $1.9\times$ the baseline's rate and faster than the standalone SLM.
Repeated generation and long-summary assessment characterize a task-dependent quality--efficiency trade-off.

We contribute a native-head/SLM division of drafting work, implemented on a mobile NPU.

\section{Related Work}
\label{sec:related}

We position \sysname\ along three related axes: where drafting executes, how draft trees are constructed, and how the verifier accepts proposed tokens.
The first two define \sysname's division of drafting work; the third determines the quality--efficiency trade-off of its default configuration.

\paragraph{Draft placement.}
In the SLM-based device--cloud systems considered here, an independent device model proposes the tokens and a server model verifies them~\citep{park2025specedge,li2025sled,li2026flexspec}.
Other work places early LLM layers on the device, as in HAT~\citep{xie2026hat}, or drafts from hidden states returned by the server~\citep{li2026efficient}.
Table~\ref{tab:rw} distinguishes these choices.
Among the listed device--cloud systems, none uses the LLM's native draft head as a mobile drafter.
Cloud+MTP uses that head on the server; \sysname\ instead executes it on the device together with an \slmdrafter.

\begin{table}[t]
  \centering
  \small
  \setlength{\tabcolsep}{3pt}
  \begin{tabular*}{\columnwidth}{@{\extracolsep{\fill}}lcccc@{}}
    \toprule
    & \textbf{D} & \textbf{H} & \textbf{S} & \textbf{V} \\
    \midrule
    SpecEdge                & \cmark  & \nomark & \nomark & \cmark  \\
    SLED                    & \cmark  & \nomark & \nomark & \cmark  \\
    HAT                     & \cmark  & \nomark & \nomark & \cmark  \\
    \citet{li2026efficient} & \cmark  & \nomark & \cmark  & \cmark  \\
    Synera                  & \cmark  & \nomark & \nomark & \nomark \\
    \midrule
    Cloud+MTP               & \nomark & \cmark  & \cmark  & \cmark  \\
    \sysname                & \cmark  & \cmark  & \cmark  & \cmark$^\dagger$  \\
    \bottomrule
  \end{tabular*}
  \caption{Comparison of draft placement and verification. \textbf{D}: device-side drafting; \textbf{H}: verifier-native draft head; \textbf{S}: server-computed hidden states or KV; \textbf{V}: token verification. For \sysname, D+H+S denotes running the native head on the device using server states. $^\dagger$Verification with $\epsilon$-threshold acceptance does not preserve the sampling distribution.}
  \label{tab:rw}
\end{table}

\paragraph{Draft construction.}
Trees provide alternative continuations for joint verification~\citep{miao2024specinfer,park2025specedge}, whereas target-alignment training and parallel-draft adaptation improve the proposal model~\citep{zhou2023distillspec,an2026pard}.
Medusa and EAGLE variants also explore hidden-state-based drafting~\citep{cai2024medusa,li2024eagle,li2025eagle3}.
\sysname\ combines a native head's chain with independent SLM alternatives.
It leaves the SLM's weights unchanged and applies QAT to the head for mobile execution (\S\ref{sec:quant}).

For an autoregressive SLM, additional depth requires sequential passes.
SpecEdge overlaps them with communication through \emph{proactive drafting} (PD), preparing a tree while verification is in flight.
\sysname\ instead obtains depth from the \mtpdrafter and uses one SLM pass for alternatives.

\paragraph{Acceptance and verification.}
Synera reduces verification frequency by retaining confident device tokens locally and sending only the rest to the server~\citep{wang2025synera}.
\sysname\ instead checks every round, but its default rule is also lossy: it can accept draft tokens that the verifier did not sample.

Verification methods differ in their acceptance criteria.
BiLD, fuzzy speculative decoding, and E$^2$-SCI use distributional divergence criteria~\citep{kim2023bild,holsman2025fuzzy,li2026e2sci}; Judge decoding uses a learned judge~\citep{bachmann2025judge}.
Medusa applies an entropy-adaptive threshold to the target's raw softmax~\citep{cai2024medusa,hewitt2022truncation}.
\sysname\ instead thresholds the verifier's sampling distribution after temperature scaling and top-$k$/top-$p$ truncation (\S\ref{sec:recovery}).
Token verification alone does not preserve selection probabilities; we therefore evaluate output quality.

\section{Method}
\label{sec:method}

Upon a user request, the cloud \verifier prefills the full prompt and computes its attention KV cache.
In parallel, the mobile \slmdrafter prefills at most the last $W$ prompt tokens and builds an initial SLM-only draft tree, since the \mtpdrafter has not yet received verifier state.
The server verifies this tree and returns accepted draft tokens, a bonus token, and the state needed to initialize the mobile head.

\begin{figure}[t]
  \centering
  \includegraphics[width=\columnwidth]{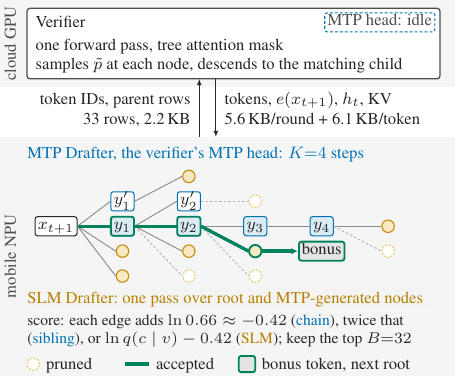}
  \caption{Hybrid draft tree in one \sysname\ round. The MTP chain $y_{1:4}$ and siblings $y'_k$ are augmented by a single SLM pass, then pruned to $B{=}32$ draft nodes plus the root. Twelve nodes are shown for clarity; the green path illustrates exact verification (default $\epsilon$ rule: \S\ref{sec:recovery}).}
  \label{fig:tree}
\end{figure}

Subsequent rounds use the \hybriddrafter (Figure~\ref{fig:tree}; Algorithm~\ref{alg:round}).
The \mtpdrafter builds a top-1 chain from the returned hidden state and KV entries, and one \slmdrafter pass adds alternatives.
The device prunes the tree to $B$ draft nodes for single-pass server verification, after which the next hand-off begins another round.
The cloud retains full-prompt context while the mobile drafters use bounded windows.
Sections~\ref{sec:tree}--\ref{sec:recovery} describe tree construction, NPU constraints, and the default lossy acceptance rule.

\subsection{Hybrid draft tree}
\label{sec:tree}

In each subsequent decoding round, the root $x_{t+1}$ is the bonus token returned by the previous verification.
The mobile \mtpdrafter receives its input embedding $e(x_{t+1})$, the preceding verifier hidden state $h_t$ from which that token was sampled, and KV entries from the verifier's last full-attention and last sliding-window layers.
The head uses this state to propose $K$ top-1 tokens sequentially.
Its inputs are transferred as 8-bit integers: 5.6\,KB per round for the input vector and approximately 6.1\,KB per committed token for the KV entries.
The server therefore returns state instead of executing the head's $K$ drafting steps; Appendix~\ref{app:qat} specifies the transferred tensors.

For the initial SLM-only round, the server's first reply seeds the \mtpdrafter's KV window with up to $W$ prompt positions.
Subsequent replies transfer KV entries only for newly committed positions, enabling the hybrid workflow without local MTP prompt prefill.

The standard configuration uses a $K{=}4$ chain and $B{=}32$ draft nodes.
At the first two chain positions, the \mtpdrafter also adds its second-ranked token as a sibling of the top-1 token, providing an alternative when that token is not accepted.
One pass of the \slmdrafter, Gemma~4 E2B from Google's QAT checkpoint~\citep{google2026gemma4e2bqat}, then proposes children under the root and each MTP-generated node.
These proposals can exceed $B$, so a common score ranks chain nodes, runner-up siblings, and SLM candidates before pruning.
\sysnametwo adds a second SLM pass over newly added nodes; its trade-off is evaluated separately.

Let $s(v)$ be the score of a parent node and let $\lambda{=}0.66$.
A chain edge adds $\ln\lambda$ to $s(v)$.
A runner-up sibling adds $2\ln\lambda$ on its own edge, placing it below the corresponding chain token.
An SLM candidate $c$ instead adds $\ln q(c\mid v)+\ln\lambda$, where $q$ is the \slmdrafter's distribution over the keep set.
Scores accumulate along the path from the root.

The value $\lambda$ summarizes the chain's measured conditional acceptance rates under exact verification: these rates are 0.65--0.68 across depths, with a geometric mean of 0.66.
Thus, $d\ln\lambda$ serves as an acceptance-based score for a depth-$d$ chain node.
Appendix~\ref{app:selection} provides the measurement setting.

The device retains the $B{=}32$ highest-scoring draft nodes with their ancestors and sends token IDs and parent indices to the server.
Including the root, verification uses 33 rows and the same CUDA-graph shape as SpecEdge-style.
The server evaluates these rows in one forward pass with a tree attention mask~\citep{miao2024specinfer}.
Under exact verification, it samples from $\tilde p(\cdot\mid v)$ at each visited node and follows the matching child.
The first sample without a matching child is committed as the bonus token and ends the round.
MAT counts both accepted draft tokens and this bonus token, which becomes the next root.
Section~\ref{sec:recovery} changes only the acceptance rule, not the work split.

This construction assigns depth to the \mtpdrafter and alternatives to the \slmdrafter.
To execute both on the NPU, \sysname\ must also fit their arithmetic, output layers, and context state to the device.
The next three subsections describe these deployment constraints; their effects on proposal acceptance are distinct from the relaxation introduced in \S\ref{sec:recovery}.

\subsection{Quantization}
\label{sec:quant}

Both drafters use static integer graphs to execute on the mobile NPU~\citep{tan2024mobilequant}, but they reach this format differently.
The released SLM QAT checkpoint has 2-, 4-, or 8-bit weights with floating-point computation~\citep{gemmateam2026gemma4,google2026gemma4e2bqat}.
We preserve its weights and convert it through graph surgery and activation calibration, without retraining.
The verifier's MTP head is released in BF16, so we apply 4-bit QAT to its linear weights.
Both models use 16-bit fixed-point activations and graphs compiled with QAIRT~\citep{qualcomm2026qairt}.

Final activation calibration and head QAT use 60 prompts from UC-tr, the \texttt{train\_sft} split of UltraChat 200k.
The head is trained for 100 steps to minimize KL divergence from the BF16 head on the verifier's greedy answers.
These prompts are separate from the UC-tr conversations used for vocabulary counts; no SB prompt is used for final calibration or QAT.
Appendix~\ref{app:qat} gives the formats, training details, and fidelity measurements.

\subsection{Restricted output vocabulary}
\label{sec:vocab}

Both drafters have output layers over 262,144 tokens.
Evaluating these layers and searching their top candidates for every SLM row and MTP step contributes substantially to mobile work.
Following frequency-based vocabulary restriction~\citep{zhao2025frspec}, we use a shared keep set of 86,089 tokens.
It contains 81,131 frequent assistant-response tokens from 20,000 UC-tr conversations, together with 24 special tokens, 256 byte tokens, and all 4,678 Hangul tokens.
Appendix~\ref{app:qat} describes this construction; Appendix~\ref{app:selection} compares keep-set sizes.

Only the drafter output vocabulary is restricted.
The verifier still samples from the full vocabulary, and a token outside the keep set can be returned as a bonus token.
The SLM retains complete input tables, while the \mtpdrafter receives the bonus token's embedding in the hand-off and uses a local 8-bit, 242\,MB table for its own drafted tokens.
Under exact verification, these restrictions may reduce acceptance but preserve the output distribution.

\subsection{1K-token context window}
\label{sec:window}

Both drafters use a $W{=}1{,}024$-token context window, following bounded-context drafting approaches~\citep{sadhukhan2025magicdec,valliappan2026windowedmtp}.
The \slmdrafter maintains its own KV ring over the latest 1,024 committed tokens, reduced to 512 in sliding-window layers, and prefills at most the last 1,024 prompt tokens.
The \mtpdrafter instead reads a window of KV entries computed by the \verifier and requires no local prompt prefill.
The \verifier itself processes the complete prompt; the mobile window does not truncate its context.

The 1K window is a device-memory and prefill-time choice.
The short UC-val prompts did not distinguish the tested lengths reliably, so these comparisons do not establish that 1K is optimal for long-context acceptance.
The tested settings appear in Appendix~\ref{app:selection}; \S\ref{sec:experiments} evaluates 32K inputs.

\subsection{\texorpdfstring{$\epsilon$}{Epsilon}-threshold acceptance}
\label{sec:recovery}

The mobile restrictions above change the proposals available for verification.
Even when the tree contains plausible continuations, exact verification ends a round if the verifier's sample matches no child.
Default \sysname\ trades exact distribution preservation for additional acceptance using an $\epsilon$-threshold rule.
A child $c$ of node $v$ is eligible if it matches the verifier's sample or if $\tilde p(c\mid v)\ge\epsilon$, where $\tilde p$ includes temperature scaling and top-$k$/top-$p$ truncation.
The fixed probability threshold follows $\epsilon$-sampling~\citep{hewitt2022truncation}.
Under a one-hot greedy distribution, the threshold admits no additional token.

Because several children can be eligible, the verifier must select one path.
It chooses the deepest path consisting entirely of eligible edges.
Ties are broken first by the number of edges matching verifier samples and then by the product of $\tilde p$ along the path.
The verifier appends its sample at the selected endpoint as the bonus token.
This is a path-level preference: a sampled edge need not be retained if another eligible path is deeper.

All committed tokens lie in the support of $\tilde p$, but their selection probabilities can differ from $\tilde p$.
The rule is therefore lossy; support membership does not imply distribution preservation or equivalent quality.
Setting $\epsilon{=}1$ recovers exact verification, and \S\ref{sec:quality-main} evaluates the quality trade-off of the default $\epsilon{=}\hyEps$ configuration.

We select $\epsilon{=}\hyEps$ on UC-val (Figure~\ref{fig:accept-rules}) using the screening procedure in \S\ref{sec:setup} and Appendix~\ref{app:selection}, and apply it at every node in the default configuration.
\sysname\ exact uses $\epsilon{=}1$, and SpecEdge-$\epsilon$ uses the same threshold and path-selection rule as \sysname.
Matching the rule does not match proposal distributions or establish equal output quality, so the experiments report both efficiency and quality.

\section{Experiments}
\label{sec:experiments}

\subsection{Setup}
\label{sec:setup}

The \verifier is Gemma~4 26B-A4B~\citep{gemmateam2026gemma4} with 4-bit activation-aware weight quantization~\citep[AWQ;][]{lin2024awq}, served on one RTX PRO 6000 cloud GPU.
A Galaxy S26 Ultra with an SM8850 chipset runs the drafters on its Hexagon Tensor Processor v81 over a commercial LTE/5G connection; the operator assigned LTE-CA or 5G NSA during measurement.
The verifier samples with temperature 1, top-$k$ 64 and top-$p$ 0.95.
Thus, exact verification preserves this sampling distribution, not a fixed greedy output.
All configurations use SpecEdge's verification server~\citep{park2025specedge}, extended with our verifier and its MTP head, the same kernels, and one client at a time\appx{ (Appendix~\ref{app:verifier})}.

\paragraph{Configuration selection.}
We compare candidate settings on UC-val, 60 prompts from UltraChat 200k's \texttt{test\_sft} split~\citep{ding2023enhancing,tunstall2024zephyr}, and fix the selected configuration before the reported SB and \infsum\ runs.
Server cost $g$ includes prompt-prefill and decoding GPU time divided by committed output tokens.
Its decode-only component equals mean GPU time per round divided by MAT.
The screening procedure minimizes $g$ among settings whose LLM-judge scores fall within the observed range of exact-verification runs.
This range is an empirical screen, not an equivalence margin.
Differences in $g$ below the approximately 2\% validation variation are resolved using the stated device-cost or comparison rationale.
Appendix~\ref{app:selection} reports the range, session-specific settings, and earlier SB development trials.

\begin{figure}[t]
  \centering
  \includegraphics[width=\columnwidth]{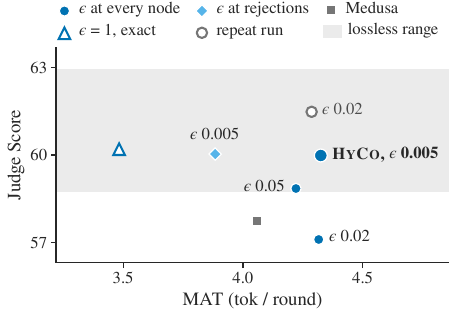}
  \caption{MAT against judge score of each acceptance rule on \sysname's tree on UC-val. The shaded band is the observed score range of exact-verification reference runs used for screening, not a quality-equivalence region. Repeat run: $\epsilon{=}0.02$ at every node, run again with node logging. Over the other runs, $g$ falls as MAT rises.}
  \label{fig:accept-rules}
\end{figure}

Figure~\ref{fig:accept-rules} shows the UC-val MAT--judge-score trade-off used to choose $\epsilon{=}\hyEps$ at every node: MAT rises while the measured judge score remains within the shaded exact-reference range. This range is a screening heuristic, not an equivalence test (Appendix~\ref{app:selection}).

\subsection{Baselines and variants}
\label{sec:baselines}

We use three baseline groups to isolate different questions.
Cloud+MTP measures how much server-side work remains after native self-speculation; SpecEdge-style and SpecEdge-$\epsilon$ test whether hybrid drafting improves on an SLM-only device tree; and \sysname\ variants expose alternative mobile--server work splits.

\emph{Cloud+MTP} uses the released BF16 MTP head over the full prefix and vocabulary, with exact verification and the same chain depth $K{=}4$ as \sysname.
\emph{SpecEdge-style} drafts a 32-node tree in four sequential SLM passes, with the same 1K mobile prefill bound.
\emph{SpecEdge-style + PD} additionally uses proactive drafting while the verification reply is in flight; both variants verify exactly.
\emph{SpecEdge-$\epsilon$} keeps this SLM-only construction but applies exactly the same threshold and path-selection rule as \sysname.
This matches the acceptance rule, not the proposal distribution or output quality, so we compare quality separately.

\sysname\ uses one SLM pass; \sysnametwo\ uses two.
Both apply $\epsilon{=}\hyEps$.
\sysname\ \emph{exact} sets $\epsilon{=}1$, which adds no token that the verifier would not sample.
The \emph{server-head} variant also verifies exactly but leaves the head on the server, in BF16 with the 86K keep set and 1K window, and omits runner-up siblings.
It is an alternative configuration, not a placement-only ablation\appx{ (Table~\ref{tab:configs})}.
\emph{On-device 8K} is the SLM alone with an 8K context, full vocabulary, and greedy decoding, evaluated on SB only; it is a standalone local reference, not a matched drafter ablation.

\begin{table*}[t]
  \centering
  \small
  \setlength{\tabcolsep}{3pt}
  \begin{tabular*}{\textwidth}{@{\extracolsep{\fill}}lcrrrrrr@{}}
\toprule
& & \multicolumn{3}{c}{SB-480} & \multicolumn{3}{c}{\infsum\ 32K} \\
\cmidrule(lr){3-5}\cmidrule(l){6-8}
Configuration & Rule & $\Delta g$ (\%) & MAT & Tok/s & $\Delta g$ (\%) & MAT & Tok/s \\
\midrule
Cloud+MTP & exact & 0.0 & 3.05 & -- & 0.0 & 2.37 & -- \\
SpecEdge-style & exact & $-20.6$ & 3.53 & 16.1 & $-2.9$ & 2.52 & 10.0 \\
SpecEdge-style + PD & exact & $-23.9$ & 3.69 & 12.1 & $-2.9$ & 2.52 & 8.2 \\
SpecEdge-$\epsilon$ & $\epsilon$ & $-29.6\pm0.2$ & 4.13 & $18.9\pm0.1$ & $-15.7$ & 3.08 & 12.4 \\
\sysname & $\epsilon$ & $-38.6\pm0.2$ & 4.71 & $35.0\pm0.7$ & $-29.2$ & 3.73 & 21.0 \\
\sysnametwo & $\epsilon$ & $-44.7$ & 5.27 & 29.0 & $-31.9$ & 3.95 & 18.8 \\
\sysname\ exact & exact & $-27.7$ & 3.84 & 28.2 & $-11.9$ & 2.81 & 16.8 \\
\sysname\ server head & exact & $-14.6$ & 3.79 & 36.1 & $-3.8$ & 2.70 & 19.8 \\
\bottomrule
\end{tabular*}

  \caption{Server cost, acceptance, and delivery speed. $\Delta g$ is the paired change from Cloud+MTP; its reference $g$ is 7.81/23.91\,ms per token on SB/32K. Lower $\Delta g$ is better. \sysname and SpecEdge-$\epsilon$ SB entries average three runs ($\pm$: run SD); all other entries use one run. MAT run SDs are both 0.012 for the repeated systems. --: Cloud+MTP was measured without the phone, so no mobile delivery rate is available. PD uses 558 SB turns and 49 summaries\appx{; recording details are in Appendix~\ref{app:sysmetrics}}. Matching rules does not enforce equal quality.}
  \label{tab:main}
\end{table*}

\subsection{Evaluation and uncertainty}
\label{sec:eval}

\paragraph{Benchmarks and quality metrics.}
We use all 480 Spec-Bench questions~\citep[SB-480;][]{xia2024unlocking} to measure decoding efficiency, including both MT-bench turns, with a 1,024-token output cap.
For SB quality, GPT-5.4 scores first-turn non-QA answers from 0 to 100 using the question but no reference answer.
This yields 398 paired questions for \sysname--Cloud+MTP; complete-case repeated comparisons involving SpecEdge-$\epsilon$ use 396.
QA tasks are evaluated separately\appx{ in Appendix~\ref{app:quality}}, and reference metrics supplement the question-only judge.
Two summarization items consistently refused by the judge's filter are omitted.

For \infsum, we evaluate the first 50 English novels in HELMET's seed-42 order, truncated to 32K tokens with its preprocessing~\citep{zhang2024infinitebench,yen2025helmet}, with a 1,200-token output cap.
We report ROUGE-L-Sum~\citep{lin2004rouge} and HELMET's model-based fluency-weighted F1, key-point recall, and precision using GPT-5.4.
Precision measures consistency with the expert summary, rather than factual accuracy against the full novel.

\paragraph{Efficiency metrics.}
We report GPU time per output token $g$, including prompt prefill; MAT; and delivery speed, defined as pooled output tokens divided by mobile wall time from request start to last token.
Mobile draft time excludes verification waits and prefill.
These metrics separate server execution, acceptance, and user-visible delivery rather than collapsing them into one speedup.

\paragraph{Repeated runs and uncertainty.}
\sysname and SpecEdge-$\epsilon$ have three SB generations each, judged once per output; other setups use one run.
The $\pm$ values are run SDs.
Paired 95\% CIs use 2,000 question bootstraps, clustering MT-bench turns.
For quality, we average scores per question before resampling\appx{ (Appendix~\ref{app:quality})}.
The CIs capture question-level uncertainty but do not separately estimate generation or judge variability.

\subsection{Efficiency under matched acceptance rules}
\label{sec:exp-main}

The strongest matched comparison uses the same $\epsilon$ rule.
\sysname\ reduces GPU time by $38.6\pm0.2$\% on SB versus $29.6\pm0.2$\% for SpecEdge-$\epsilon$, an additional 9.0 percentage points.
Delivery speeds are $35.0\pm0.7$ and $18.9\pm0.1$ tokens/s, respectively.
The hybrid tree achieves MAT 4.71 with one SLM pass, whereas the SLM-only tree reaches 4.13 with four.
On 32K summarization, their reductions are 29.2\% (95\% CI: 27.8--30.6\%) and 15.7\% (14.1--17.1\%), at 21.0 and 12.4 tokens/s.

The design also helps under exact verification.
On SB, \sysname\ exact reduces $g$ by 27.7\%, compared with 20.6\% for SpecEdge-style exact.
On 32K, the corresponding savings are 11.9\% and 2.9\%.
The exact and same-$\epsilon$ comparisons therefore provide complementary evidence: the hybrid drafting design saves server work without changing the verifier's sampling distribution, and an additional advantage remains when both device drafters use the same relaxed rule.

\subsection{Quality and failure cases}
\label{sec:quality-main}

The following measurements characterize the observed quality trade-off of the lossy default; they are not equivalence tests.
The three run-level judge means are $90.34\pm0.50$ for \sysname and $90.60\pm0.61$ for SpecEdge-$\epsilon$.
Their paired difference on common questions is $-0.2$ points (95\% CI: $-1.2$ to $+0.8$).
Relative to Cloud+MTP, \sysname differs by $+0.2$ points ($-0.9$ to $+1.3$).
Against \sysname\ exact, its three-run per-question mean differs by $-0.4$ points ($-1.5$ to $+0.6$), using the same 398 questions and one exact run.
Intervals spanning zero do not establish quality equivalence.

Translation is an important exception to the aggregate pattern: \sysname scores 2.6 points below SpecEdge-$\epsilon$ (95\% CI: $-5.2$ to $-0.2$), while its difference from Cloud+MTP spans zero.
SpecEdge-$\epsilon$ is higher in all three translation runs.
The 15 task-wise contrasts are exploratory and unadjusted for multiple comparisons; we do not claim that every task is unaffected\appx{ (Table~\ref{tab:quality-ci})}.

\paragraph{Reference-based SB diagnostics.}
For \sysname versus Cloud+MTP, GSM8K numeric-answer accuracy is 95.0\% versus 95.0\%, RAG contains-gold accuracy is 65.7\% versus 67.1\%, CNN/DailyMail ROUGE-L is 19.2 versus 19.8, and WMT14 corpus BLEU is 33.9 versus 33.7.
These checks use \sysname's first SB generation and are descriptive single-generation comparisons, not three-run averages; RAG measures normalized reference containment, not full answer correctness\appx{ (Appendix~\ref{app:quality})}.

\paragraph{Long-summary diagnostics.}
On 32K summaries, \sysname's HELMET F1 difference from Cloud+MTP is $+0.4$ points ($-2.3$ to $+3.1$), and its precision difference is $+1.9$ ($-2.6$ to $+6.2$).
ROUGE-L-Sum is 37.5 versus 38.5, a difference of $-1.0$ ($-2.1$ to $+0.1$).
SpecEdge-$\epsilon$ instead combines higher precision with lower recall\appx{ (Table~\ref{tab:long-quality})}.
These reference-based measures add evidence about summary content but are not a guarantee of factual consistency.

A fixed $\pm20$-point screen of \sysname's three-run per-question means against Cloud+MTP selects 13 decreases and 12 increases.
Our manual inspection identifies six substantive errors among the decreases, as well as score/content discrepancies in both directions.
We retain all measured scores; the selected cases are not an estimate of the overall error rate\appx{ (Appendix~\ref{app:failures})}.

\subsection{Mobile costs and head transfer}
\label{sec:analysis}

\paragraph{Mobile work and delivery.}
On SB, \sysname delivers \hyTestTokSB\ tokens/s versus \hyOnDeviceTok\ for the SLM alone.
Its \hyDevTokHyco\,ms of mobile draft time per output token is about a third of the SLM alone's \hyDevTokSlm\,ms.
SpecEdge-style exact delivers 16.1 tokens/s and consumes 42.8\,ms of mobile draft time per token.
The two efficiency goals therefore use distinct references: less server work than Cloud+MTP, and faster delivery than the local SLM.
Since Cloud+MTP was not streamed to the phone, we do not compare its user delivery speed with \sysname.

\paragraph{Head transfer and fidelity.}
In a separate paired 150-question experiment, the device-head configuration (W4, 86K keep set, 1K window) has \hyTestHeadLoss\% lower MAT than the original BF16 server head (95\% CI: \hyTestHeadLossLo--\hyTestHeadLossHi\%).
In the full SB experiment, relaxing \sysname exact raises MAT from 3.84 to 4.71.
These are separate comparisons, not additive estimates of independent component effects.

\paragraph{Startup latency.}
The first hand-off introduces a startup cost.
Single-run \sysname traces have mean time to first token of 846\,ms on SB and 4.60\,s on 32K inputs.
Thus, higher average delivery speed does not imply a lower first-token delay.
We report GPU-busy efficiency, not loaded-service capacity or energy savings\appx{ (Appendix~\ref{app:sysmetrics})}.

\section{Discussion}
\label{sec:discussion}

\subsection{From server accelerator to client drafter}
\label{sec:disc-protocol}

\sysname's central systems change is to turn a server-internal self-speculation component into a device--cloud protocol.
Device--cloud systems commonly offload an independent draft model~\citep{park2025specedge,li2025sled,li2026flexspec}, while a server with a native draft head can already self-speculate~\citep{lyu2026distance}.
\sysname\ moves that head's sequential drafting steps to the mobile device and returns the slice of verifier state the head reads.
The independent SLM then supplies alternative branches without taking over the chain's depth.

This split is most useful when server GPU time is constrained, the head is small enough for the device, and its inputs fit the link.
Our measurements establish the resulting single-client execution-time trade-off, not the capacity gain of a loaded service.
Appendix~\ref{app:selection} records the development-stage server/mobile exchange that motivated the placement choice.

\subsection{Choosing the mobile--cloud work split}
\label{sec:disc-split}

The measured configurations span several resource axes: server GPU time, mobile drafting work and delivery speed, first-token latency, and an unmeasured energy cost.
Prior systems adapt draft length to the channel~\citep{li2026flexspec}, an acceptance threshold~\citep{li2026e2sci}, or which tokens reach the server~\citep{wang2025synera}.
Here, the choices also include where the native head executes and how many SLM passes accompany it.
Figure~\ref{fig:overview} illustrates example deployment priorities rather than an evaluated online switching policy.

The variants occupy different points in this space.
A second SLM pass in \sysnametwo\ increases server saving but lowers delivery speed relative to default \sysname.
On long summaries, its additional server saving is small because prefill---which drafting does not shorten---accounts for about a fifth of $g$.
Moving the head back to the server reduces mobile draft time but gives up server-side saving.
Cloud+MTP removes mobile drafting altogether.
Thus, no setting dominates every resource, and timing measurements alone do not determine an energy-optimal choice.

\subsection{Lossy acceptance revisited}
\label{sec:disc-lossy}

\citet{wang2026revisiting} identify a general risk of truncation-based verification: restricting which tokens can be accepted does not preserve the verifier's relative preferences within that set.
Our \epsacc belongs to this family.
The repeated SB generations and 32K summary evaluation in \S\ref{sec:quality-main} characterize its observed trade-off on the evaluated tasks; they do not remove the underlying distributional risk.

Two comparisons separate the drafting design from the relaxed rule.
Under exact verification, \sysname\ outperforms the exact SLM-only baseline on both SB and 32K inputs.
Under the same $\epsilon$ rule, it also retains an efficiency advantage over SpecEdge-$\epsilon$.
Matching the acceptance rule does not match the proposal distribution, and the translation result shows that task-specific quality differences can remain.
The fixed-tree replay in Appendix~\ref{app:selection} helps explain which proposal sources gain acceptance, but it is mechanistic evidence rather than a quality guarantee for newly generated answers.

\section{Conclusion}
\label{sec:conclusion}

\sysname\ pairs the verifier's native MTP head for draft depth with one SLM pass for alternative branches, moving sequential head steps to the device without repeated SLM passes for depth.

Relative to Cloud+MTP, exact verification reduces server GPU time per token by 27.7\% on SB and 11.9\% on 32K-token summaries.
Default \sysname\ uses \epsacc to increase these savings to \hyTestSaveSB\% and \hyTestSaveHm\%, with task-dependent quality--efficiency trade-offs evaluated through repeated SB generations and long-summary metrics.
SB delivery also exceeds standalone local generation.

These single-client results do not establish a universal serving optimum.
Full cross-family deployment needs an on-device native head; Appendix~\ref{app:crossfamily} shows SLM transfer to Muse-Glimmer-30B via a token bridge without retraining.

\section*{Limitations}
\label{sec:limitations}

\paragraph{Deployment scope and cost.}
Our measurements establish single-client execution efficiency, not loaded-service capacity or end-to-end monetary or energy efficiency.
Each round waits for one network round trip, which bounds mobile delivery speed, and the full hybrid deployment uses one mobile device, one commercial cellular network, one cloud GPU type, and one verifier family.
All configurations serve one client at a time in our engine.
Retaining unverified device tokens, as in Synera~\citep{wang2025synera}, would introduce a different lossy mechanism.

\paragraph{Distribution and task coverage.}
Default \sysname\ does not preserve the verifier's sampling distribution.
The reported judge-score interval permits a modest quality decrease, and translation favors SpecEdge-$\epsilon$ in an exploratory comparison.
The aggregate judge result covers scored non-QA first-turn answers, not QA or second-turn quality.
The question-only judge and reference-summary-based HELMET metrics do not comprehensively assess factual consistency, and we do not establish how relaxed acceptance changes output diversity.

\paragraph{Statistical scope.}
Only the two main relaxed systems have three SB generations; the other configurations and Cloud+MTP have one.
Question-bootstrap intervals condition on these recorded outputs, and one judge evaluation per output does not separate judge variability from generation variability.
The UC-val screening range is not an equivalence margin.
\appx{Appendices~\ref{app:qat} and~\ref{app:selection} distinguish calibration and development data from the reported benchmark runs.}

\paragraph{Requirements and device costs.}
\sysname\ requires access to the verifier's native head, intermediate hidden states, and KV entries, which may not be available from closed model services.
The head must also fit the mobile device, and the SLM requires a shared tokenizer or token bridge.
Heads that use more verifier layers~\citep{chen2026dflash} may require larger state transfers.
The \mtpdrafter adds about 20\,ms per mobile round and up to 6.3\,MB to the first reply of a long prompt, and the device stores 365\,MB for it alongside the SLM's 754\,MB graph \unskip{}.

These implementation requirements also constrain where the measured acceptance rates may transfer.
The UC-derived keep set and 1K window need not suit every language or domain~\citep{abramovich2026speedbench}.
Tokens outside the keep set comprise 3.0\% of verifier bonus tokens on UC-val and 6.8\% on the 150-question SB subset of the head-configuration comparison (\S\ref{sec:analysis}) \unskip{}.
Exact verification preserves the output distribution despite these proposal restrictions; relaxed acceptance does not.
The SLM accesses earlier context only indirectly through the MTP chain.
We evaluate text only, not Gemma's image or audio inputs.

\paragraph{Information exposure.}
Each client holds the MTP head and part of the verifier's input embeddings.
Transmitted activations and reply-size patterns may expose information beyond the text~\citep{cunningham2026privacy}.
We evaluate no safeguard.

\bibliography{refs}

@inproceedings{moon2024frontier,
  title         = {A New Frontier of {AI}: On-Device {AI} Training and Personalization},
  author        = {Moon, Jijoong and Lee, Hyeonseok and Chu, Jiho and Park, Donghak and Hong, Seungbaek and Seo, Hyungjun and others},
  booktitle     = {Proceedings of the 46th International Conference on Software Engineering: Software Engineering in Practice (ICSE-SEIP)},
  pages         = {323--333},
  year          = {2024},
  publisher     = {Association for Computing Machinery},
  doi           = {10.1145/3639477.3639716}
}

@inproceedings{leviathan2023fast,
  title         = {Fast Inference from Transformers via Speculative Decoding},
  author        = {Leviathan, Yaniv and Kalman, Matan and Matias, Yossi},
  booktitle     = {Proceedings of the 40th International Conference on Machine Learning (ICML)},
  series        = {Proceedings of Machine Learning Research},
  volume        = {202},
  pages         = {19274--19286},
  year          = {2023},
  publisher     = {PMLR},
  url           = {https://proceedings.mlr.press/v202/leviathan23a.html}
}

@misc{chen2023accelerating,
  title         = {Accelerating Large Language Model Decoding with Speculative Sampling},
  author        = {Chen, Charlie and Borgeaud, Sebastian and Irving, Geoffrey and Lespiau, Jean-Baptiste and Sifre, Laurent and Jumper, John},
  year          = {2023},
  eprint        = {2302.01318},
  archivePrefix = {arXiv},
  primaryClass  = {cs.CL},
  doi           = {10.48550/arXiv.2302.01318},
  url           = {https://arxiv.org/abs/2302.01318}
}

@inproceedings{xia2024unlocking,
  title         = {Unlocking Efficiency in Large Language Model Inference: A Comprehensive Survey of Speculative Decoding},
  author        = {Xia, Heming and Yang, Zhe and Dong, Qingxiu and Wang, Peiyi and Li, Yongqi and Ge, Tao and others},
  booktitle     = {Findings of the Association for Computational Linguistics: ACL 2024},
  pages         = {7655--7671},
  year          = {2024},
  publisher     = {Association for Computational Linguistics},
  doi           = {10.18653/v1/2024.findings-acl.456},
  url           = {https://aclanthology.org/2024.findings-acl.456/}
}

@inproceedings{yen2025helmet,
  title         = {{HELMET}: How to Evaluate Long-context Models Effectively and Thoroughly},
  author        = {Yen, Howard and Gao, Tianyu and Hou, Minmin and Ding, Ke and Fleischer, Daniel and Izsak, Peter and others},
  booktitle     = {International Conference on Learning Representations (ICLR)},
  pages         = {98914--98965},
  year          = {2025},
  url           = {https://proceedings.iclr.cc/paper_files/paper/2025/hash/f5332c8273d02729730a9c24dec2135e-Abstract-Conference.html}
}

@inproceedings{zhang2024infinitebench,
  title         = {$\infty${Bench}: Extending Long Context Evaluation Beyond {100K} Tokens},
  author        = {Zhang, Xinrong and Chen, Yingfa and Hu, Shengding and Xu, Zihang and Chen, Junhao and Hao, Moo Khai and others},
  booktitle     = {Proceedings of the 62nd Annual Meeting of the Association for Computational Linguistics (Volume 1: Long Papers)},
  pages         = {15262--15277},
  year          = {2024},
  publisher     = {Association for Computational Linguistics},
  doi           = {10.18653/v1/2024.acl-long.814},
  url           = {https://aclanthology.org/2024.acl-long.814/}
}

@inproceedings{lin2004rouge,
  title         = {{ROUGE}: A Package for Automatic Evaluation of Summaries},
  author        = {Lin, Chin-Yew},
  booktitle     = {Text Summarization Branches Out},
  pages         = {74--81},
  year          = {2004},
  address       = {Barcelona, Spain},
  publisher     = {Association for Computational Linguistics},
  url           = {https://aclanthology.org/W04-1013/}
}

@inproceedings{lin2024awq,
  title         = {{AWQ}: Activation-aware Weight Quantization for On-Device {LLM} Compression and Acceleration},
  author        = {Lin, Ji and Tang, Jiaming and Tang, Haotian and Yang, Shang and Chen, Wei-Ming and Wang, Wei-Chen and others},
  booktitle     = {Proceedings of Machine Learning and Systems (MLSys)},
  volume        = {6},
  pages         = {87--100},
  year          = {2024},
  url           = {https://proceedings.mlsys.org/paper_files/paper/2024/hash/42a452cbafa9dd64e9ba4aa95cc1ef21-Abstract-Conference.html}
}

@inproceedings{cai2024medusa,
  title         = {Medusa: Simple {LLM} Inference Acceleration Framework with Multiple Decoding Heads},
  author        = {Cai, Tianle and Li, Yuhong and Geng, Zhengyang and Peng, Hongwu and Lee, Jason D. and Chen, Deming and others},
  booktitle     = {Proceedings of the 41st International Conference on Machine Learning (ICML)},
  series        = {Proceedings of Machine Learning Research},
  volume        = {235},
  pages         = {5209--5235},
  year          = {2024},
  publisher     = {PMLR},
  url           = {https://proceedings.mlr.press/v235/cai24b.html}
}

@inproceedings{li2024eagle,
  title         = {{EAGLE}: Speculative Sampling Requires Rethinking Feature Uncertainty},
  author        = {Li, Yuhui and Wei, Fangyun and Zhang, Chao and Zhang, Hongyang},
  booktitle     = {Proceedings of the 41st International Conference on Machine Learning (ICML)},
  series        = {Proceedings of Machine Learning Research},
  volume        = {235},
  pages         = {28935--28948},
  year          = {2024},
  publisher     = {PMLR},
  url           = {https://proceedings.mlr.press/v235/li24bt.html}
}

@misc{deepseekai2024v3,
  title         = {{DeepSeek-V3} Technical Report},
  author        = {{DeepSeek-AI}},
  year          = {2024},
  eprint        = {2412.19437},
  archivePrefix = {arXiv},
  primaryClass  = {cs.CL},
  doi           = {10.48550/arXiv.2412.19437},
  url           = {https://arxiv.org/abs/2412.19437}
}

@misc{liu2024turbospec,
  title         = {{TurboSpec}: Closed-loop Speculation Control System for Optimizing {LLM} Serving Goodput},
  author        = {Liu, Xiaoxuan and Park, Jongseok and Hu, Langxiang and Kwon, Woosuk and Li, Zhuohan and Zhang, Chen and others},
  year          = {2024},
  eprint        = {2406.14066},
  archivePrefix = {arXiv},
  primaryClass  = {cs.AI},
  doi           = {10.48550/arXiv.2406.14066},
  url           = {https://arxiv.org/abs/2406.14066},
  note          = {Earlier version titled ``Optimizing Speculative Decoding for Serving Large Language Models Using Goodput''}
}

@inproceedings{park2025specedge,
  title         = {{SpecEdge}: Scalable Edge-Assisted Serving Framework for Interactive {LLMs}},
  author        = {Park, Jinwoo and Cho, Seunggeun and Han, Dongsu},
  booktitle     = {Advances in Neural Information Processing Systems (NeurIPS)},
  volume        = {38},
  year          = {2025},
  doi           = {10.52202/085713-3101},
  url           = {https://proceedings.neurips.cc/paper_files/paper/2025/hash/8587069d00a69d0ea498d547fffad6dd-Abstract-Conference.html}
}

@inproceedings{li2025sled,
  title         = {{SLED}: A Speculative {LLM} Decoding Framework for Efficient Edge Serving},
  author        = {Li, Xiangchen and Spatharakis, Dimitrios and Ghafouri, Saeid and Fan, Jiakun and Vandierendonck, Hans and John, Deepu and others},
  booktitle     = {Proceedings of the Tenth ACM/IEEE Symposium on Edge Computing (SEC)},
  pages         = {1--8},
  year          = {2025},
  publisher     = {Association for Computing Machinery},
  doi           = {10.1145/3769102.3770608}
}

@misc{lyu2026distance,
  title         = {Speculation at a Distance: Where Edge-Cloud Speculative Decoding Actually Pays Off},
  author        = {Lyu, Yuan and Irukulapati, Bharath and Champati, Jaya Prakash},
  year          = {2026},
  eprint        = {2606.25091},
  archivePrefix = {arXiv},
  primaryClass  = {cs.DC},
  doi           = {10.48550/arXiv.2606.25091},
  url           = {https://arxiv.org/abs/2606.25091}
}

@inproceedings{miao2024specinfer,
  title         = {{SpecInfer}: Accelerating Large Language Model Serving with Tree-based Speculative Inference and Verification},
  author        = {Miao, Xupeng and Oliaro, Gabriele and Zhang, Zhihao and Cheng, Xinhao and Wang, Zeyu and Zhang, Zhengxin and others},
  booktitle     = {Proceedings of the 29th ACM International Conference on Architectural Support for Programming Languages and Operating Systems, Volume 3 (ASPLOS)},
  pages         = {932--949},
  year          = {2024},
  publisher     = {Association for Computing Machinery},
  doi           = {10.1145/3620666.3651335}
}

@inproceedings{sadhukhan2025magicdec,
  title         = {{MagicDec}: Breaking the Latency-Throughput Tradeoff for Long Context Generation with Speculative Decoding},
  author        = {Sadhukhan, Ranajoy and Chen, Jian and Chen, Zhuoming and Tiwari, Vashisth and Lai, Ruihang and Shi, Jinyuan and others},
  booktitle     = {International Conference on Learning Representations (ICLR)},
  pages         = {6835--6850},
  year          = {2025},
  url           = {https://proceedings.iclr.cc/paper_files/paper/2025/hash/13f972adf12bdf886583d48cd528002f-Abstract-Conference.html}
}

@inproceedings{chen2026dflash,
  title         = {{DFlash}: Block Diffusion for Flash Speculative Decoding},
  author        = {Chen, Jian and Liang, Yesheng and Liu, Zhijian},
  booktitle     = {Proceedings of the 43rd International Conference on Machine Learning (ICML)},
  series        = {Proceedings of Machine Learning Research},
  volume        = {306},
  pages         = {15343--15355},
  year          = {2026},
  publisher     = {PMLR},
  url           = {https://proceedings.mlr.press/v306/chen26bv.html}
}

@article{li2026flexspec,
  title         = {{FlexSpec}: Frozen Drafts Meet Evolving Targets in Edge-Cloud Collaborative {LLM} Speculative Decoding},
  author        = {Li, Yuchen and Kong, Rui and Lyu, Zhonghao and Li, Qiyang and Chen, Xinran and Cai, Hengyi and others},
  journal       = {IEEE Transactions on Mobile Computing},
  volume        = {25},
  number        = {9},
  pages         = {14642--14655},
  year          = {2026},
  publisher     = {IEEE},
  doi           = {10.1109/TMC.2026.3683574}
}

@inproceedings{li2025eagle3,
  title         = {{EAGLE-3}: Scaling up Inference Acceleration of Large Language Models via Training-Time Test},
  author        = {Li, Yuhui and Wei, Fangyun and Zhang, Chao and Zhang, Hongyang},
  booktitle     = {Advances in Neural Information Processing Systems (NeurIPS)},
  volume        = {38},
  year          = {2025},
  doi           = {10.52202/085713-4562},
  url           = {https://proceedings.neurips.cc/paper_files/paper/2025/hash/c7b5a35ea98b62512a869c19ea7b03cb-Abstract-Conference.html}
}

@inproceedings{abramovich2026speedbench,
  title         = {{SPEED-Bench}: A Unified and Diverse Benchmark for Speculative Decoding},
  author        = {Abramovich, Talor and Ashkenazi, Maor and Putterman, Izzy and Chislett, Benjamin and Mitra, Tiyasa and Rouhani, Bita Darvish and others},
  booktitle     = {Proceedings of the 43rd International Conference on Machine Learning (ICML)},
  series        = {Proceedings of Machine Learning Research},
  volume        = {306},
  pages         = {240--267},
  year          = {2026},
  publisher     = {PMLR},
  url           = {https://proceedings.mlr.press/v306/abramovich26a.html}
}

@inproceedings{bachmann2025judge,
  title         = {Judge Decoding: Faster Speculative Sampling Requires Going Beyond Model Alignment},
  author        = {Bachmann, Gregor and Anagnostidis, Sotiris and Pumarola, Albert and Georgopoulos, Markos and Sanakoyeu, Artsiom and Du, Yuming and others},
  booktitle     = {International Conference on Learning Representations (ICLR)},
  pages         = {40894--40914},
  year          = {2025},
  url           = {https://proceedings.iclr.cc/paper_files/paper/2025/hash/656d6174fd5ffb01c843f269649ab5cb-Abstract-Conference.html}
}

@inproceedings{zhou2023distillspec,
  title         = {{DistillSpec}: Improving Speculative Decoding via Knowledge Distillation},
  author        = {Zhou, Yongchao and Lyu, Kaifeng and Rawat, Ankit Singh and Menon, Aditya Krishna and Rostamizadeh, Afshin and Kumar, Sanjiv and others},
  booktitle     = {International Conference on Learning Representations (ICLR)},
  pages         = {32011--32050},
  year          = {2024},
  url           = {https://proceedings.iclr.cc/paper_files/paper/2024/hash/8766fbc68e1ed1cdef712ce273e0a363-Abstract-Conference.html}
}

@inproceedings{an2026pard,
  title         = {{PARD}: Accelerating {LLM} Inference with Low-Cost {PAR}allel Draft Model Adaptation},
  author        = {An, Zihao and Bai, Huajun and Liu, Ziqiong and Li, Dong and Barsoum, Emad},
  booktitle     = {International Conference on Learning Representations (ICLR)},
  year          = {2026},
  url           = {https://openreview.net/forum?id=XbOyv7iVGL}
}

@inproceedings{liu2023llm,
  title         = {{LLM-QAT}: Data-Free Quantization Aware Training for Large Language Models},
  author        = {Liu, Zechun and Oguz, Barlas and Zhao, Changsheng and Chang, Ernie and Stock, Pierre and Mehdad, Yashar and others},
  booktitle     = {Findings of the Association for Computational Linguistics: ACL 2024},
  pages         = {467--484},
  year          = {2024},
  publisher     = {Association for Computational Linguistics},
  doi           = {10.18653/v1/2024.findings-acl.26},
  url           = {https://aclanthology.org/2024.findings-acl.26/}
}

@misc{gemmateam2026gemma4,
  title         = {{Gemma 4} Technical Report},
  author        = {{Gemma Team}},
  year          = {2026},
  eprint        = {2607.02770},
  archivePrefix = {arXiv},
  primaryClass  = {cs.CL},
  doi           = {10.48550/arXiv.2607.02770},
  url           = {https://arxiv.org/abs/2607.02770}
}

@misc{google2026gemma4e2bqat,
  title         = {{gemma-4-E2B-it-qat-mobile-transformers}},
  author        = {{Google DeepMind}},
  year          = {2026},
  howpublished  = {Hugging Face model card},
  url           = {https://huggingface.co/google/gemma-4-E2B-it-qat-mobile-transformers}
}

@misc{gemmateam2025gemma,
  title         = {{Gemma 3} Technical Report},
  author        = {{Gemma Team}},
  year          = {2025},
  eprint        = {2503.19786},
  archivePrefix = {arXiv},
  primaryClass  = {cs.CL},
  doi           = {10.48550/arXiv.2503.19786},
  url           = {https://arxiv.org/abs/2503.19786}
}

@misc{bengio2013estimating,
  title         = {Estimating or Propagating Gradients Through Stochastic Neurons for Conditional Computation},
  author        = {Bengio, Yoshua and L{\'e}onard, Nicholas and Courville, Aaron},
  year          = {2013},
  eprint        = {1308.3432},
  archivePrefix = {arXiv},
  primaryClass  = {cs.LG},
  doi           = {10.48550/arXiv.1308.3432},
  url           = {https://arxiv.org/abs/1308.3432}
}

@inproceedings{ding2023enhancing,
  title         = {Enhancing Chat Language Models by Scaling High-quality Instructional Conversations},
  author        = {Ding, Ning and Chen, Yulin and Xu, Bokai and Qin, Yujia and Hu, Shengding and Liu, Zhiyuan and others},
  booktitle     = {Proceedings of the 2023 Conference on Empirical Methods in Natural Language Processing},
  pages         = {3029--3051},
  year          = {2023},
  publisher     = {Association for Computational Linguistics},
  doi           = {10.18653/v1/2023.emnlp-main.183},
  url           = {https://aclanthology.org/2023.emnlp-main.183/}
}

@inproceedings{tunstall2024zephyr,
  title         = {Zephyr: Direct Distillation of {LM} Alignment},
  author        = {Tunstall, Lewis and Beeching, Edward Emanuel and Lambert, Nathan and Rajani, Nazneen and Rasul, Kashif and Belkada, Younes and others},
  booktitle     = {First Conference on Language Modeling (COLM)},
  year          = {2024},
  url           = {https://openreview.net/forum?id=aKkAwZB6JV}
}

@inproceedings{zhao2025frspec,
  title         = {{FR-Spec}: Accelerating Large-Vocabulary Language Models via Frequency-Ranked Speculative Sampling},
  author        = {Zhao, Weilin and Pan, Tengyu and Han, Xu and Zhang, Yudi and Sun, Ao and Huang, Yuxiang and others},
  booktitle     = {Proceedings of the 63rd Annual Meeting of the Association for Computational Linguistics (Volume 1: Long Papers)},
  pages         = {3909--3921},
  year          = {2025},
  publisher     = {Association for Computational Linguistics},
  doi           = {10.18653/v1/2025.acl-long.198},
  url           = {https://aclanthology.org/2025.acl-long.198/}
}

@misc{valliappan2026windowedmtp,
  title         = {{Windowed-MTP}: Removing the Full-Context Draft-{KV} Tax at Million-Token Context},
  author        = {Valliappan, Alagappan},
  year          = {2026},
  eprint        = {2607.21535},
  archivePrefix = {arXiv},
  primaryClass  = {cs.LG},
  doi           = {10.48550/arXiv.2607.21535},
  url           = {https://arxiv.org/abs/2607.21535}
}

@misc{li2026efficient,
  title         = {Efficient and Privacy Aware Edge Cloud Collaborative Inference for Large Language Models},
  author        = {Li, Cheng and Liu, Jiexiong and Chen, Yixuan and Li, Yi},
  year          = {2026},
  eprint        = {2607.13093},
  archivePrefix = {arXiv},
  primaryClass  = {cs.CR},
  doi           = {10.48550/arXiv.2607.13093},
  url           = {https://arxiv.org/abs/2607.13093}
}

@inproceedings{holsman2025fuzzy,
  title         = {Fuzzy Speculative Decoding for a Tunable Accuracy-Runtime Tradeoff},
  author        = {Holsman, Maximilian and Huang, Yukun and Dhingra, Bhuwan},
  booktitle     = {Findings of the Association for Computational Linguistics: ACL 2025},
  pages         = {26257--26273},
  year          = {2025},
  publisher     = {Association for Computational Linguistics},
  doi           = {10.18653/v1/2025.findings-acl.1346},
  url           = {https://aclanthology.org/2025.findings-acl.1346/}
}

@inproceedings{hewitt2022truncation,
  title         = {Truncation Sampling as Language Model Desmoothing},
  author        = {Hewitt, John and Manning, Christopher D. and Liang, Percy},
  booktitle     = {Findings of the Association for Computational Linguistics: EMNLP 2022},
  pages         = {3414--3427},
  year          = {2022},
  publisher     = {Association for Computational Linguistics},
  doi           = {10.18653/v1/2022.findings-emnlp.249},
  url           = {https://aclanthology.org/2022.findings-emnlp.249/}
}

@misc{wang2025synera,
  title         = {Synera: Synergistic {LLM} Serving across Device and Cloud at Scale},
  author        = {Wang, Genglin and Zeng, Liekang and Yang, Bufang and Liu, Kaiwei and Xing, Guoliang and Sun, Chumin and others},
  year          = {2025},
  eprint        = {2511.07423},
  archivePrefix = {arXiv},
  primaryClass  = {cs.DC},
  doi           = {10.48550/arXiv.2511.07423},
  url           = {https://arxiv.org/abs/2511.07423}
}

@misc{wang2026revisiting,
  title         = {Revisiting Lossy Verification in Speculative Decoding: Mechanisms, Trade-offs, and Failure Modes},
  author        = {Wang, Tianyu and Zhou, Yuxuan and Li, Heng and Wang, Wenbin and Xiao, Zikai and Zheng, Chunrui and others},
  year          = {2026},
  eprint        = {2607.26627},
  archivePrefix = {arXiv},
  primaryClass  = {cs.CL},
  doi           = {10.48550/arXiv.2607.26627},
  url           = {https://arxiv.org/abs/2607.26627}
}

@inproceedings{li2026e2sci,
  title         = {{E$^2$-SCI}: Elastic Edge-Cloud Speculative Decoding via Credit Inertia},
  author        = {Li, Senyao and Wang, Haozhao and Jiang, Zhaobai and Jin, Zhanbo and Fan, Hao and Li, Ruixuan},
  booktitle     = {Proceedings of the IEEE/CVF Conference on Computer Vision and Pattern Recognition (CVPR)},
  pages         = {12944--12954},
  year          = {2026},
  url           = {https://openaccess.thecvf.com/content/CVPR2026/html/Li_E2-SCI_Elastic_Edge-Cloud_Speculative_Decoding_via_Credit_Inertia_CVPR_2026_paper.html}
}

@inproceedings{xie2026hat,
  title         = {A Novel Hat-Shaped Device-Cloud Collaborative Inference Framework for Large Language Models},
  author        = {Xie, Zuan and Xu, Yang and Xu, Hongli and Liao, Yunming and Yao, Zhiwei},
  booktitle     = {IEEE INFOCOM 2026 -- IEEE Conference on Computer Communications},
  pages         = {1--10},
  year          = {2026},
  publisher     = {IEEE},
  doi           = {10.1109/INFOCOM59046.2026.11571521}
}

@misc{cunningham2026privacy,
  title         = {Privacy-Aware Split Inference with Speculative Decoding for Large Language Models over Wide-Area Networks},
  author        = {Cunningham, Michael},
  year          = {2026},
  eprint        = {2602.16760},
  archivePrefix = {arXiv},
  primaryClass  = {cs.CR},
  doi           = {10.48550/arXiv.2602.16760},
  url           = {https://arxiv.org/abs/2602.16760}
}

@inproceedings{tan2024mobilequant,
  title         = {{MobileQuant}: Mobile-friendly Quantization for On-device Language Models},
  author        = {Tan, Fuwen and Lee, Royson and Dudziak, {\L}ukasz and Hu, Shell Xu and Bhattacharya, Sourav and Hospedales, Timothy and others},
  booktitle     = {Findings of the Association for Computational Linguistics: EMNLP 2024},
  pages         = {9761--9771},
  year          = {2024},
  publisher     = {Association for Computational Linguistics},
  doi           = {10.18653/v1/2024.findings-emnlp.570},
  url           = {https://aclanthology.org/2024.findings-emnlp.570/}
}

@inproceedings{kim2023bild,
  title         = {Speculative Decoding with Big Little Decoder},
  author        = {Kim, Sehoon and Mangalam, Karttikeya and Moon, Suhong and Malik, Jitendra and Mahoney, Michael W. and Gholami, Amir and others},
  booktitle     = {Advances in Neural Information Processing Systems (NeurIPS)},
  volume        = {36},
  pages         = {39236--39256},
  year          = {2023},
  doi           = {10.52202/075280-1705},
  url           = {https://proceedings.neurips.cc/paper_files/paper/2023/hash/7b97adeafa1c51cf65263459ca9d0d7c-Abstract-Conference.html}
}

@inproceedings{zheng2023judging,
  title         = {Judging {LLM}-as-a-Judge with {MT-Bench} and {Chatbot Arena}},
  author        = {Zheng, Lianmin and Chiang, Wei-Lin and Sheng, Ying and Zhuang, Siyuan and Wu, Zhanghao and Zhuang, Yonghao and others},
  booktitle     = {Advances in Neural Information Processing Systems (NeurIPS), Datasets and Benchmarks Track},
  volume        = {36},
  pages         = {46595--46623},
  year          = {2023},
  doi           = {10.52202/075280-2020},
  url           = {https://proceedings.neurips.cc/paper_files/paper/2023/hash/91f18a1287b398d378ef22505bf41832-Abstract-Datasets_and_Benchmarks.html}
}

@inproceedings{bojar2014findings,
  title         = {Findings of the 2014 Workshop on Statistical Machine Translation},
  author        = {Bojar, Ond{\v{r}}ej and Buck, Christian and Federmann, Christian and Haddow, Barry and Koehn, Philipp and Leveling, Johannes and others},
  booktitle     = {Proceedings of the Ninth Workshop on Statistical Machine Translation},
  pages         = {12--58},
  year          = {2014},
  publisher     = {Association for Computational Linguistics},
  doi           = {10.3115/v1/W14-3302},
  url           = {https://aclanthology.org/W14-3302/}
}

@inproceedings{nallapati2016abstractive,
  title         = {Abstractive Text Summarization using Sequence-to-sequence {RNN}s and Beyond},
  author        = {Nallapati, Ramesh and Zhou, Bowen and dos Santos, Cicero and G{\"u}l{\c{c}}ehre, {\c{C}}a{\u{g}}lar and Xiang, Bing},
  booktitle     = {Proceedings of the 20th {SIGNLL} Conference on Computational Natural Language Learning},
  pages         = {280--290},
  year          = {2016},
  publisher     = {Association for Computational Linguistics},
  doi           = {10.18653/v1/K16-1028},
  url           = {https://aclanthology.org/K16-1028/}
}

@article{kwiatkowski2019natural,
  title         = {{Natural Questions}: A Benchmark for Question Answering Research},
  author        = {Kwiatkowski, Tom and Palomaki, Jennimaria and Redfield, Olivia and Collins, Michael and Parikh, Ankur and Alberti, Chris and others},
  journal       = {Transactions of the Association for Computational Linguistics},
  volume        = {7},
  pages         = {452--466},
  year          = {2019},
  doi           = {10.1162/tacl_a_00276},
  url           = {https://aclanthology.org/Q19-1026/}
}

@inproceedings{karpukhin2020dense,
  title         = {Dense Passage Retrieval for Open-Domain Question Answering},
  author        = {Karpukhin, Vladimir and Oguz, Barlas and Min, Sewon and Lewis, Patrick and Wu, Ledell and Edunov, Sergey and others},
  booktitle     = {Proceedings of the 2020 Conference on Empirical Methods in Natural Language Processing (EMNLP)},
  pages         = {6769--6781},
  year          = {2020},
  publisher     = {Association for Computational Linguistics},
  doi           = {10.18653/v1/2020.emnlp-main.550},
  url           = {https://aclanthology.org/2020.emnlp-main.550/}
}

@misc{cobbe2021training,
  title         = {Training Verifiers to Solve Math Word Problems},
  author        = {Cobbe, Karl and Kosaraju, Vineet and Bavarian, Mohammad and Chen, Mark and Jun, Heewoo and Kaiser, Lukasz and others},
  year          = {2021},
  eprint        = {2110.14168},
  archivePrefix = {arXiv},
  primaryClass  = {cs.LG},
  doi           = {10.48550/arXiv.2110.14168},
  url           = {https://arxiv.org/abs/2110.14168}
}

@inproceedings{paszke2019pytorch,
  title         = {{PyTorch}: An Imperative Style, High-Performance Deep Learning Library},
  author        = {Paszke, Adam and Gross, Sam and Massa, Francisco and Lerer, Adam and Bradbury, James and Chanan, Gregory and others},
  booktitle     = {Advances in Neural Information Processing Systems (NeurIPS)},
  volume        = {32},
  year          = {2019},
  url           = {https://proceedings.neurips.cc/paper_files/paper/2019/hash/bdbca288fee7f92f2bfa9f7012727740-Abstract.html}
}

@inproceedings{kwon2023efficient,
  title         = {Efficient Memory Management for Large Language Model Serving with {PagedAttention}},
  author        = {Kwon, Woosuk and Li, Zhuohan and Zhuang, Siyuan and Sheng, Ying and Zheng, Lianmin and Yu, Cody Hao and others},
  booktitle     = {Proceedings of the 29th Symposium on Operating Systems Principles (SOSP)},
  pages         = {611--626},
  year          = {2023},
  publisher     = {Association for Computing Machinery},
  doi           = {10.1145/3600006.3613165}
}

@inproceedings{frantar2025marlin,
  title         = {{MARLIN}: Mixed-Precision Auto-Regressive Parallel Inference on Large Language Models},
  author        = {Frantar, Elias and Castro, Roberto L. and Chen, Jiale and Hoefler, Torsten and Alistarh, Dan},
  booktitle     = {Proceedings of the 30th ACM SIGPLAN Annual Symposium on Principles and Practice of Parallel Programming (PPoPP)},
  pages         = {239--251},
  year          = {2025},
  publisher     = {Association for Computing Machinery},
  doi           = {10.1145/3710848.3710871}
}

@inproceedings{ye2025flashinfer,
  title         = {{FlashInfer}: Efficient and Customizable Attention Engine for {LLM} Inference Serving},
  author        = {Ye, Zihao and Chen, Lequn and Lai, Ruihang and Lin, Wuwei and Zhang, Yineng and Wang, Stephanie and others},
  booktitle     = {Proceedings of Machine Learning and Systems (MLSys)},
  volume        = {7},
  year          = {2025},
  url           = {https://proceedings.mlsys.org/paper_files/paper/2025/hash/dbf02b21d77409a2db30e56866a8ab3a-Abstract-Conference.html}
}

@inproceedings{zheng2024sglang,
  title         = {{SGLang}: Efficient Execution of Structured Language Model Programs},
  author        = {Zheng, Lianmin and Yin, Liangsheng and Xie, Zhiqiang and Sun, Chuyue and Huang, Jeff and Yu, Cody Hao and others},
  booktitle     = {Advances in Neural Information Processing Systems (NeurIPS)},
  volume        = {37},
  pages         = {62557--62583},
  year          = {2024},
  doi           = {10.52202/079017-2000}
}

@inproceedings{wolf2020transformers,
  title         = {Transformers: State-of-the-Art Natural Language Processing},
  author        = {Wolf, Thomas and Debut, Lysandre and Sanh, Victor and Chaumond, Julien and Delangue, Clement and Moi, Anthony and others},
  booktitle     = {Proceedings of the 2020 Conference on Empirical Methods in Natural Language Processing: System Demonstrations},
  pages         = {38--45},
  year          = {2020},
  publisher     = {Association for Computational Linguistics},
  doi           = {10.18653/v1/2020.emnlp-demos.6},
  url           = {https://aclanthology.org/2020.emnlp-demos.6/}
}

@misc{openai2026gpt54,
  title         = {{GPT-5.4 Thinking} System Card},
  author        = {{OpenAI}},
  year          = {2026},
  howpublished  = {\url{https://deploymentsafety.openai.com/gpt-5-4-thinking}},
  url           = {https://deploymentsafety.openai.com/gpt-5-4-thinking},
  note          = {Accessed 2026-10-08}
}

@misc{meta2026museglimmer,
  title         = {{Muse-Glimmer-30B}},
  author        = {{Meta Superintelligence Lab}},
  year          = {2026},
  howpublished  = {Hugging Face model card},
  url           = {https://huggingface.co/meta-models/Muse-Glimmer-30B}
}

@misc{nvidia2026museglimmernvfp4,
  title         = {{Muse-Glimmer-30B-NVFP4}},
  author        = {{NVIDIA}},
  year          = {2026},
  howpublished  = {Hugging Face model card},
  url           = {https://huggingface.co/nvidia/Muse-Glimmer-30B-NVFP4}
}

@misc{incoai2026museglimmerdflash2,
  title         = {{Muse-Glimmer-30B-DFlash2}},
  author        = {{Inco AI}},
  year          = {2026},
  howpublished  = {Hugging Face model card},
  url           = {https://huggingface.co/incoai/Muse-Glimmer-30B-DFlash2}
}

@misc{qualcomm2026qairt,
  title         = {{Qualcomm AI Runtime (QAIRT) SDK}},
  author        = {{Qualcomm Technologies, Inc.}},
  year          = {2026},
  howpublished  = {\url{https://softwarecenter.qualcomm.com/catalog/item/Qualcomm_AI_Runtime_SDK}},
  url           = {https://softwarecenter.qualcomm.com/catalog/item/Qualcomm_AI_Runtime_SDK},
  note          = {Version 2.47.1. Accessed 2026-10-08}
}

\ifwithappendix
\clearpage
\appendix
\section{The Drafters on the Mobile Device}
\label{app:qat}
\suppressfloats[t]

Table~\ref{tab:formats} shows each drafter's integer formats, size and training.
Weights are per-channel symmetric integers at the bit maps it lists, activation ranges are fixed ahead of time, and no operator falls back to floating point.
QAIRT compiles static shapes, with 32 tree rows for the \slmdrafter and one query position over a 1,024-column window for the \mtpdrafter.

\begin{table}[t]
  \centering
  \small
  \setlength{\tabcolsep}{2.5pt}
  \begin{tabular}{@{}L{0.22\columnwidth}L{0.34\columnwidth}L{0.38\columnwidth}@{}}
    \toprule
    & \slmdrafter & \mtpdrafter \\
    \midrule
    Weight bit map   & the checkpoint's 2/4/8 bits & W4 (W8: post-training) \\
    Activations      & 16-bit                & 16-bit \\
    KV / inputs      & 8-bit KV cache        & 16-bit (8-bit on the wire) \\
    Embeddings       & 2/4-bit LUT           & 8-bit LUT, 242\,MB$^\ast$ \\
    Size             & 754\,MB NPU graph     & 123\,MB NPU graph at W4 \\
    Our training     & none (graph surgery only) & QAT, 100 steps (W4) \\
    \bottomrule
  \end{tabular}
  \caption{Integer formats of the two drafters on the NPU. W4, W8: 4- and 8-bit weights. LUT: lookup table on the CPU. $^\ast$For the \mtpdrafter's own drafted tokens.}
  \label{tab:formats}
\end{table}

\paragraph{Final calibration and training.}
The collaborative configurations in Table~\ref{tab:main} use the UC-tr-calibrated SLM and, where applicable, MTP head.
Both calibration procedures use 60 first-turn prompts from the training split of UltraChat 200k and the verifier's greedy answers; no SB prompt is used in this calibration or QAT.
The keep-set token counts use separate UC-tr conversations.
Validation and benchmark configurations are distinguished below.

The graph surgery of the \slmdrafter keeps the checkpoint's bit allocation and the weights' integers and scales bit-exact, rewrites each low-bit weight matrix multiplication as an integer one with an explicit encoding of its bit width, decomposes root-mean-square (RMS) normalization into standard operators, computes the rotary tables on the host, and keeps the embedding tables on the CPU as lookup tables \unskip{}.
For the final calibration, activation ranges come from the 60 UC-tr prompts with the verifier's greedy answers used for the MTP head's QAT, and KV caches use 8 bits with a range of 1.2 times their 99.9th percentile \unskip{}.
Keys enter the KV ring with their rotary embedding applied, so an entry keeps its absolute position wherever it sits in the ring \unskip{}.

Table~\ref{tab:qat-hparams} shows the UC-tr training settings for the MTP head, whose KL loss follows Gemma's own QAT~\citep{gemmateam2025gemma} and data-free QAT for LLMs~\citep{liu2023llm}.
After a per-channel search for the clipping ratio that minimizes the quantization error, every linear layer is fake-quantized with a straight-through estimator~\citep{bengio2013estimating}.
The training inputs, captured by teacher forcing the \verifier in BF16, are the input embedding, the last hidden state, and the keys and values of the two source layers \unskip{}.
The 60 UC-tr prompts are first turns sampled at random from conversations that the keep set's counts never read.
The top-1 agreement of \S\ref{sec:quant} compares the quantized integer head with the BF16 head on the 25,625 response tokens of the 60 UC-val sequences.
The trained integers and scales form a static graph whose output layer is untied and restricted to the keep set, and whose fixed input ranges are the 99.99th percentiles of the UC-tr captures.

\begin{table}[t]
  \centering
  \small
  \setlength{\tabcolsep}{3pt}
  \begin{tabular}{@{}L{0.31\columnwidth}L{0.65\columnwidth}@{}}
    \toprule
    Initialization   & BF16 release \\
    Teacher          & original head (BF16) \\
    Trained          & linear weights only (norms frozen) \\
    Weight bits      & 4, all linear layers incl.\ output layer \\
    Weight clipping  & per-channel mean-squared-error search, ratio 1.0--0.6 \\
    Loss             & KL on response tokens \\
    Optimizer        & AdamW, $\beta{=}(0.9, 0.95)$, no weight decay, gradient-norm clip 1.0 \\
    Learning rate    & $10^{-5}$, 50 warm-up steps, cosine \\
    Batch            & 1 sequence, 2 accumulation steps \\
    Sequence length  & $\le$2,048 (about 900) \\
    Steps            & 100 \\
    Data             & 60 UC-tr prompts with the BF16 \verifier's greedy answers ($\le$1,024 new tokens), all for training, and the 60 UC-val prompts, answered the same way ($\le$512 new tokens), for evaluation \\
    Precision        & fp32 master weights, BF16 autocast \\
    \bottomrule
  \end{tabular}
  \caption{Final UC-tr QAT of the MTP head's 4-bit weights.}
  \label{tab:qat-hparams}
\end{table}

The keep set always holds the 24 special tokens, the 256 byte tokens, which can spell any string, and all 4,678 Hangul tokens regardless of their UC-tr frequency \unskip{}.
The other 81,131 are the tokens most frequent in the assistant turns of 20,000 UC-tr conversations.
The \slmdrafter's input tables stay complete, and the \mtpdrafter's first step reads $e(x_{t+1})$ from the hand-off, so both drafters can take a committed token outside the keep set.

The hand-off carries the \mtpdrafter's inputs as 8-bit integers \unskip{}.
These are $e(x_{t+1})$ and $h_t$, the latter taken before the \verifier's final normalization, 5,632 values per round, and the keys and values of the last full-attention and the last sliding-window layer, 6,144 values per committed token.
The first reply seeds the \mtpdrafter's window with the prompt's last 1,024 tokens, at most 6.3\,MB \unskip{}.

\section{Verifier Engine}
\label{app:verifier}

We ported the \verifier's text model with static shapes into SpecEdge's server~\citep{park2025specedge}, replacing its model side, and checked it against the Hugging Face model \unskip{}.
Attention projections run as 4-bit integer Marlin matrix products~\citep{frantar2025marlin} and the experts, 8 of 128 per token, through vLLM's fused Marlin mixture-of-experts kernel~\citep{kwon2023efficient}, while the dense MLP and the router stay in 16 bits.
RMS normalization runs in FlashInfer~\citep{ye2025flashinfer}, and attention in PyTorch~\citep{paszke2019pytorch} with the tree mask.
Tree verification uses the routed MoE kernel described above rather than a dense all-expert path.
Each decode round is padded to the narrowest of ten attention windows that covers its context, 256, 512 and 768 positions and then powers of two up to the maximum context of 65,536, and replays that window's CUDA graphs, while prefill runs eagerly in chunks \unskip{}.
The \verifier's MTP head, which Cloud+MTP and the \serverhead variant run on the server, replays its own graphs over the same windows, and its chains agree with eager execution on 99.2--99.6\% of tokens, as often as two stock attention kernels agree.

\section{Decoding Round and Configurations}
\label{app:configs}

Algorithms~\ref{alg:round}--\ref{alg:verify} specify one \sysname decoding round as implemented.
Algorithm~\ref{alg:round} is the round, and it calls the \mtpdrafter (Algorithm~\ref{alg:mtp}), the construction of the hybrid draft tree (Algorithm~\ref{alg:tree}) and the verification (Algorithm~\ref{alg:verify}).
Every \sysname configuration uses chain depth $K{=}4$, runner-up siblings at depths up to $D{=}2$, node budget $B{=}32$, depth cost $\ln\lambda$ with $\lambda{=}0.66$, $W{=}1{,}024$-token windows, the keep set $\mathcal V$ of 86,089 tokens and 32 \slmdrafter candidates per row, and $\epsilon{=}\hyEps$ except in the exact variant.
Positions count from 0, so the root $x_{t+1}$ sits at position $t$, and $\tilde p$ applies temperature 1, top-$k$ 64 and top-$p$ 0.95.

\begin{algorithm}[t]
  \small
  \caption{One \sysname decoding round.}
  \label{alg:round}
  \begin{algorithmic}[1]
    \Require root $x_{t+1}$; hand-off $H$ of the last verification
    \Statex \emph{Mobile device}
    \State $(y_{1:K},y'_{1:D})\gets\Call{MtpDraft}{H}$ \Comment{Alg.~\ref{alg:mtp}}
    \State $\mathcal T\gets\Call{BuildTree}{x_{t+1},y_{1:K},y'_{1:D}}$; send $\mathcal T$ \Comment{Alg.~\ref{alg:tree}}
    \Statex \emph{Server}
    \State $(\hat y,b,H)\gets\Call{Verify}{\mathcal T}$; reply with them \Comment{Alg.~\ref{alg:verify}}
    \Statex \emph{Mobile device}
    \State \label{alg:walk}$P\gets$ deepest path below $x_{t+1}$ whose every node $c$ with parent $v$ has $c=\hat y_v$; ties to the lowest row
    \State commit the tokens of $P$ and the bonus token $b$
    \State move the \slmdrafter KV rows of $P$'s nodes behind $x_{t+1}$'s; the next \slmdrafter pass writes those of $b$ and of $P$'s nodes without one
    \State $t\gets t+1+|P|$; \ $x_{t+1}\gets b$
  \end{algorithmic}
\end{algorithm}

\begin{algorithm}[t]
  \small
  \caption{\mtpdrafter on the mobile device, \textsc{MtpDraft}.}
  \label{alg:mtp}
  \begin{algorithmic}[1]
    \Require $H=(t,a,\hat x,R)$: root position $t$, $\hat x=[e(x_{t+1});h_t]$, and $R$, the KV rows of positions $[a,t)$ in the \verifier's last full-attention and last sliding-window layers, all 8-bit
    \Statex \textbf{State:} per layer a ring of $W$ KV columns; live positions $[\ell,t_0)$
    \If{$a\ne t_0$} clear the ring; $\ell\gets a$ \Comment{new request}
    \EndIf
    \State write the row of position $p$ into column $p\bmod W$, for $a\le p<t$, rescaled to the graph's 16-bit inputs; $t_0\gets t$
    \State $\ell\gets\max(\ell,t-W)$; mask all columns outside $[\ell,t)$
    \State fix the query's rotary position at $t-1$; the keys arrive rotated
    \State $z\gets\hat x$
    \For{$k=1,\dots,K$}
      \State $(o,h)\gets$ head graph on $z$ over the ring, which it does not extend \Comment{$o$ over $\mathcal V$}
      \State $i,i'\gets$ indices of the largest and second-largest entry of $o$, first among equals
      \State $y_k\gets\mathcal V[i]$; \ if $k\le D$, $y'_k\gets\mathcal V[i']$
      \State $z\gets[E[i];h]$ \Comment{$E$: 8-bit table of $e(\cdot)$ over $\mathcal V$}
    \EndFor
    \State \Return $y_{1:K}$, $y'_{1:D}$
  \end{algorithmic}
\end{algorithm}

\begin{algorithm}[t]
  \small
  \caption{Hybrid draft tree on the mobile device, \textsc{BuildTree}.}
  \label{alg:tree}
  \begin{algorithmic}[1]
    \Require root $r$; chain $y_{1:K}$; runner-ups $y'_{1:D}$
    \State $\mathcal T\gets\{r\}$; \ $s(r)\gets0$; \ $u_0\gets r$
    \For{$k=1,\dots,K$} \Comment{chain}
      \State add $y_k$ under $u_{k-1}$ with $s(y_k)=s(u_{k-1})+\ln\lambda$; \ $u_k\gets y_k$
    \EndFor
    \For{$k=1,\dots,D$} \Comment{runner-up siblings}
      \State if $y'_k\ne y_k$, add $y'_k$ under $u_{k-1}$ with $s(y'_k)=s(u_{k-1})+2\ln\lambda$
    \EndFor
    \State \label{alg:tree-f}$F\gets$ committed tokens without \slmdrafter KV, ending with $r$, and the MTP-generated nodes, by descending $s$
    \State one \slmdrafter forward over $F$; row $v$ attends to the last $W$ committed positions and to $v$'s ancestors
    \State $C\gets\emptyset$
    \ForAll{$v\in F$ except committed tokens before $r$}
      \State $q(\cdot\mid v)\gets$ softmax of row $v$'s logits over $\mathcal V$
      \ForAll{the 32 tokens $c$ of largest $q(c\mid v)$}
        \State unless $v$ has a child $c$, add $(v,c)$ to $C$ with $s(c)=s(v)+\ln q(c\mid v)+\ln\lambda$
      \EndFor
    \EndFor
    \State \label{alg:tree-admit}$\theta\gets$ $B$-th highest $s$ over $\mathcal T\setminus\{r\}$ and $C$ ($-\infty$ if fewer); add every $(v,c)\in C$ with $s(c)\ge\theta$ to $\mathcal T$
    \State \sysnametwo: lines~\ref{alg:tree-f}--\ref{alg:tree-admit} once more, with $F$ the nodes just added
    \State \label{alg:trim}keep the $B$ highest-scoring nodes, ties to the earlier one, and their ancestors
    \State \Return rows of $r$ and the kept nodes in insertion order: token ID, depth and parent row, with the ancestor mask, padded to $B{+}1$ rows by repeating the last
  \end{algorithmic}
\end{algorithm}

\begin{algorithm}[t]
  \small
  \caption{Verification on the server, \textsc{Verify}.}
  \label{alg:verify}
  \begin{algorithmic}[1]
    \Require tree $\mathcal T$ with root $r=x_{t+1}$; the \verifier's KV of positions $[0,t)$
    \State row $j$ takes position $t+\mathrm{depth}(j)$ and attends to $[0,t)$, its ancestor rows and itself \Comment{tree attention mask}
    \State one forward over the $B{+}1$ rows gives logits $o_j$ and last hidden states $h_j$, taken before the final norm
    \ForAll{rows $j$}
      \State $\tilde p(\cdot\mid j)\gets$ softmax over the 64 largest $o_j$, cut to the top-$p$ nucleus and renormalized; sample $\hat y_j\sim\tilde p(\cdot\mid j)$
    \EndFor
    \State \label{alg:accept}for each node $c$ with parent $v$: $\mathrm{hit}(c)\gets[c=\hat y_v]$; \ $\mathrm{ok}(c)\gets\mathrm{hit}(c)\lor\tilde p(c\mid v)\ge\epsilon$
    \State $A\gets$ nodes whose path from $r$ is ok at every node
    \If{$A=\emptyset$} $v^\ast\gets r$
    \Else
      \State $v^\ast\gets$ node of $A$ maximizing, in order, depth, hits on its path and $\prod\tilde p$ along it
      \State $\hat y_v\gets c$ for every edge $(v,c)$ on the path to $v^\ast$ \Comment{so line~\ref{alg:walk} of Alg.~\ref{alg:round} ends at $v^\ast$}
    \EndIf
    \State $b\gets\hat y_{v^\ast}$; \ $P\gets$ path from $r$ to $v^\ast$, without $r$
    \State move the KV of $r$ and $P$ to positions $t,\dots,t{+}|P|$ and drop the other rows; \ $t'\gets t+1+|P|$ \Comment{$b$ has no KV yet}
    \State $H\gets(t',a,[e(b);h_{v^\ast}],R)$ with $R$ the two layers' KV of positions $[a,t')$, $a=\max(t,t'-W)$, or $\max(0,t'-W)$ in a request's first round, at 8 bits with fixed scales: 44\,B and 5,632\,B per round plus 6,144\,B per position
    \State \Return $\hat y$, $b$, $H$
  \end{algorithmic}
\end{algorithm}

At a request's start, the mobile device fills the \slmdrafter's cache with the prompt's last $W$ tokens and drafts the first tree with the \slmdrafter alone, while the server prefills the whole prompt; its first $H$ thus carries up to $W$ rows, and later ones carry only the newly committed positions.
Since $\tilde p(c\mid v)\ge1$ only when $\tilde p(\cdot\mid v)$ puts all its mass on $c$, $\epsilon{=}1$ reduces line~\ref{alg:accept} of Algorithm~\ref{alg:verify} to exact verification.

Table~\ref{tab:configs} shows the settings of every configuration in Tables~\ref{tab:main} and~\ref{tab:quality}.

\begin{table*}[t]
  \centering
  \small
  \setlength{\tabcolsep}{4pt}
  \newcommand{\ind}[1]{\hspace{0.9em}#1}
  \begin{tabular}{@{}l@{\hspace{8pt}}llll@{\hspace{8pt}}clccl@{}}
    \toprule
    & \multicolumn{4}{c}{MTP head} & & & & & \\
    \cmidrule(lr){2-5}
    Configuration & Place & Weights & Vocabulary & Window & $K$ & Siblings & SLM passes & Nodes / rows & Acceptance \\
    \midrule
    Cloud+MTP         & server & BF16 & 262K & whole prefix & 4 & -- & -- & 4 / 5 & exact \\
    SpecEdge-style    & -- & -- & -- & -- & 0 & -- & 4 & 32 / 33 & exact \\
    \ind{+ PD}         & -- & -- & -- & -- & 0 & -- & 4 + PD & 32 / 33 & exact \\
    SpecEdge-$\epsilon$ & -- & -- & -- & -- & 0 & -- & 4 & 32 / 33 & $\epsilon{=}\hyEps$ \\
    \sysname          & mobile & W4 & 86K & 1,024 & 4 & depths 1--2 & 1 & 32 / 33 & $\epsilon{=}\hyEps$ \\
    \ind{-2P}         & mobile & W4 & 86K & 1,024 & 4 & depths 1--2 & 2 & 32 / 33 & $\epsilon{=}\hyEps$ \\
    \ind{exact}       & mobile & W4 & 86K & 1,024 & 4 & depths 1--2 & 1 & 32 / 33 & exact ($\epsilon{=}1$) \\
    \ind{\serverhead} & server & BF16 & 86K & 1,024 & 4 & -- & 1 & 32 / 33 & exact \\
    \midrule
    On-device 8K      & \multicolumn{9}{l}{the SLM alone: 8,192-token context, 262K vocabulary, greedy decoding, no \verifier} \\
    \bottomrule
  \end{tabular}
  \caption{Settings of every configuration in Tables~\ref{tab:main} and~\ref{tab:quality}. Every tree drafted on the mobile device uses the \slmdrafter with the 86K keep set and a 1,024-token context. $K$: MTP chain depth; Siblings: depths of the runner-up siblings; Nodes / rows: draft nodes after the trim and \verifier rows per round, including the root. $\epsilon$: \epsacc at every node (\S\ref{sec:recovery}). PD continues drafting while a verification reply is in flight. All SLM-based device configurations use the final UC-tr activation calibration.}
  \label{tab:configs}
\end{table*}

The algorithms above define the final executable decoding configuration. The next section records how its fixed parameters were selected and how the finalized runs were measured.

\section{Validation Protocol and Measurement Conditions}
\label{app:selection}

\paragraph{Final protocol.}
The configuration comparisons use UC-val on the cloud GPU and include prefill in $g$, as the benchmark runs do.
UC-val has 60 prompts, about 100 tokens at the median, and a 512-token output limit.
Earlier SB trials informed implementation development and were rechecked on UC-val.
The final 86K keep set is derived from UC-tr, and $\epsilon$ was chosen on UC-val before the final benchmark runs.
The development-stage sweeps below compare their listed configurations within each block; they are not repetitions of the final benchmark configuration.

\paragraph{Tree and acceptance-rule selection.}
The tree configuration was also screened on UC-val.
With exact verification and the mobile \mtpdrafter, the chain's per-depth conditional acceptance is 0.65--0.68, with geometric mean 0.66 over 7,513 rounds; this value sets the depth score in \S\ref{sec:tree}.
A fully accepted $K{=}4$ chain is extended to a fifth token by an SLM leaf in 17\% of rounds.
The cumulative chain acceptance reaches 18.9\% at depth 4, and later runner-up siblings add little additional server saving.
The 32-node budget matches SpecEdge-style, and a second SLM pass is retained only as the \sysnametwo\ operating option.

A separate diagnostic compares the MTP head on the device and server under the development configuration.
Moving the head to the mobile device shortens the server's median round from 26.8 to 21.7\,ms and lowers $g$ by 15.2\%, at about 20\,ms more mobile work per round.
This is not a placement-only comparison because the two configurations differ in other head settings; the final matched settings are listed in Table~\ref{tab:configs}.

Figure~\ref{fig:accept-rules} in the main text summarizes the acceptance-rule sweep.
At $\epsilon{=}\hyEps$, applying the threshold at every node rather than only at rejections lowers $g$ from 5.62 to 5.10\,ms at the same judge score of 60.0, while Medusa's typical acceptance at its defaults falls below the exact-reference score range.
We choose $\epsilon{=}\hyEps$ at every node by the screening rule.
In that session, $\epsilon{=}0.02$ at every node scored 57.1 without node logging and 61.5 with every tree node logged.
On the fixed logged trees of one run, lowering $\epsilon$ has little remaining headroom: $\epsilon{=}\hyEps$ gives MAT 4.364 versus 4.366 if every drafted token in the verifier's sampling support were accepted.

\begin{table}[t]
  \centering
  \small
  \setlength{\tabcolsep}{2.4pt}
  \begin{tabular*}{\columnwidth}{@{\extracolsep{\fill}}lrrr@{}}
    \toprule
    Setting & MAT & $g$ & Device \\
    \midrule
    \multicolumn{4}{@{}l}{\emph{Keep set}} \\
    21K                                 & 2.98 & 7.00 & 45.7 \\
    45K                                 & 3.25 & 6.44 & 47.8 \\
    86K \checkmark                      & 3.36 & 6.23 & 52.9 \\
    262K, full vocabulary               & 2.85 & 7.57 & 77.9 \\
    \multicolumn{4}{@{}l}{\emph{SLM context}} \\
    1,024 \checkmark                    & 3.08 & 6.82 & 48.2 \\
    4,096                               & 3.09 & 6.77 & 48.7 \\

    \multicolumn{4}{@{}l}{\emph{MTP head window, head on the server}} \\
    512                                 & 3.20 & 7.75 & 28.8 \\
    1,024 \checkmark                    & 3.18 & 7.83 & 28.8 \\
    2,048                               & 3.16 & 7.89 & 28.1 \\
    \bottomrule
  \end{tabular*}
  \caption{Selection runs on UC-val for the keep set, the \slmdrafter's context, and the MTP head's window. Blocks ran in different sessions and setups, so compare rows only within a block. \checkmark: chosen; $g$: server ms per output token; Device: on-device ms per round.}
  \label{tab:sel-device}
\end{table}

The SLM integer-graph diagnostic found that 16-bit floating-point activations lengthened a 32-row pass from 15.7 to 38--41\,ms at unchanged acceptance.
The final W4 MTP-head QAT raises top-1 agreement with BF16 from 87.4\% to 91.0\% on UC-val; W8 post-training quantization retains 98.8\% without training but doubles weight storage (Appendix~\ref{app:qat}).
For the vocabulary sweep, the full vocabulary lengthens mobile time per round from 52.9 to 77.9\,ms and has MAT 2.85; Table~\ref{tab:sel-device} reports the corresponding $g$ values.
On UC-val, 3.0\% of verifier bonus tokens lie outside the 86K keep set, although UC-val comes from the same dataset family used for the keep-set counts.
The SLM-window and MTP-head-window sweeps are also shown in Table~\ref{tab:sel-device}; the short prompts do not establish long-context optimality.

\paragraph{Development sweeps.}
These runs, including those of Table~\ref{tab:sel-device} for the keep set and the two window lengths, took place in different sessions and setups \unskip{}.
The runs on the siblings, the keep set and the \slmdrafter's context used the development-stage 45K keep set, which the keep-set block varies, and exact verification \unskip{}.
The head-window block kept the head on the server, restricted to a 55K superset of the 45K set \unskip{}.
The runs on the rule, Medusa's setting, the threshold and the passes ran with the 86K keep set and runner-up siblings at depths 1--2 in one session \unskip{}.
The exact-reference screening range is 58.7--63.0: 58.7 is the Cloud+MTP score in an earlier session without the mobile device, and exact \sysname\ scored 61.8, 63.0, and 60.2 in three sessions.
This range is used only for screening and is not an equivalence margin.

The quantization follows the NPU and the released checkpoints, so no UC-val run chose it.
The score penalty of \S\ref{sec:tree} is measured rather than chosen.
It is $\ln 0.66 \approx -0.42$, where 0.66 is the geometric mean of the chain's per-depth acceptance rates in \S\ref{sec:tree}, measured over 7,513 rounds with exact verification and the \mtpdrafter on the mobile device \unskip{}.

\paragraph{Test-run accounting.}
In the test runs, $g$ includes the prompt's prefill, which configurations with a mobile tree run as a separate step before their first round, and counts Cloud+MTP's tokens up to its first stop token, as the mobile device does \unskip{}.

On \infsum, the 32K setting uses the first 50 novels of HELMET's seed-42 order with its 2-shot prompt, and its prompts have a median of 30.6K Gemma tokens \unskip{}.
Answers are sampled, not greedily decoded as in HELMET.
During the test runs, the mobile device keeps its screen on at minimum brightness. Temperature was not controlled or used to adjust the reported full-run timings.
With the screen off, each round trip took 14\,ms longer \unskip{}.

\section{Quality per Task}
\label{app:quality}

\paragraph{Scoring and missing observations.}
The SB judge sees the question and answer, each truncated to 6,000 characters by our scoring script, but not a reference answer.
We retain the original scores and do not replace scores after manual inspection.
Two summarization prompts are consistently blocked by the judge's filter.
\sysname has 398 eligible scored first turns in each run; SpecEdge-$\epsilon$ has 397, 397, and 398.
The latter's three-run complete-case intersection has 396 questions.
Consequently, \sysname--Cloud+MTP uses 398 paired questions, while both repeated comparisons involving SpecEdge-$\epsilon$ use 396.
The final PD run provides 558 of 560 SB turns (479 questions), with 398 eligible first-turn judge scores, and 49 of 50 long-summary requests and scores.
A failed score or incomplete response is not imputed.

\begin{table*}[t]
  \centering\small\setlength{\tabcolsep}{4pt}
  \begin{tabular*}{\textwidth}{@{\extracolsep{\fill}}lcrrr@{}}
\toprule
Configuration & Generations & Judge mean & Paired $\Delta$ vs.\ Cloud+MTP & Paired $n$ \\
\midrule
Cloud+MTP & 1 & 90.15 & -- & -- \\
SpecEdge-style & 1 & 90.21 & $+0.1$ $[-1.3, +1.4]$ & 398 \\
SpecEdge-style + PD & 1 & 90.05 & $-0.1$ $[-1.6, +1.2]$ & 398 \\
SpecEdge-$\epsilon$ & 3 & $90.60\pm0.61$ & $+0.5$ $[-0.7, +1.8]$ & 396 \\
\sysname & 3 & $90.34\pm0.50$ & $+0.2$ $[-0.9, +1.3]$ & 398 \\
\sysnametwo & 1 & 90.38 & $+0.2$ $[-1.0, +1.5]$ & 398 \\
\sysname\ exact & 1 & 90.76 & $+0.6$ $[-0.7, +1.9]$ & 398 \\
\sysname\ server head & 1 & 90.72 & $+0.6$ $[-0.4, +1.6]$ & 398 \\
On-device 8K & 1 & 54.59 & $-35.6$ $[-39.2, -32.0]$ & 398 \\
\bottomrule
\end{tabular*}

  \caption{SB judge results. For the repeated systems, the judge mean is the mean $\pm$ sample SD of the three run-level means, each using its available eligible answers. Paired differences instead use per-question three-run averages on the common complete-case set. Thus, a paired difference need not equal the difference of the displayed marginal means. Other rows use one generation and one judge score per output. All intervals are paired question-bootstrap 95\% CIs.}
  \label{tab:quality}
\end{table*}

\begin{table*}[t]
  \centering\footnotesize\setlength{\tabcolsep}{3pt}
  \begin{tabular*}{\textwidth}{@{\extracolsep{\fill}}lrrrrrr@{}}
\toprule
Comparison & All & MT-bench & Translation & Summarization & Math & RAG \\
\midrule
\sysname $-$ SpecEdge-$\epsilon$ & \begin{tabular}[t]{@{}r@{}}$-0.2$\\$[-1.2, +0.8]$\end{tabular} & \begin{tabular}[t]{@{}r@{}}$+0.2$\\$[-1.5, +1.8]$\end{tabular} & \begin{tabular}[t]{@{}r@{}}$-2.6$\\$[-5.2, -0.2]$\end{tabular} & \begin{tabular}[t]{@{}r@{}}$+0.3$\\$[-0.1, +0.7]$\end{tabular} & \begin{tabular}[t]{@{}r@{}}$+0.5$\\$[-1.3, +2.7]$\end{tabular} & \begin{tabular}[t]{@{}r@{}}$+0.6$\\$[-2.7, +3.7]$\end{tabular} \\[2pt]
\sysname $-$ Cloud+MTP & \begin{tabular}[t]{@{}r@{}}$+0.2$\\$[-0.9, +1.3]$\end{tabular} & \begin{tabular}[t]{@{}r@{}}$-0.2$\\$[-1.9, +1.4]$\end{tabular} & \begin{tabular}[t]{@{}r@{}}$+0.3$\\$[-2.6, +3.3]$\end{tabular} & \begin{tabular}[t]{@{}r@{}}$+0.0$\\$[-0.3, +0.4]$\end{tabular} & \begin{tabular}[t]{@{}r@{}}$-0.1$\\$[-1.6, +1.9]$\end{tabular} & \begin{tabular}[t]{@{}r@{}}$+0.9$\\$[-2.9, +4.9]$\end{tabular} \\[2pt]
SpecEdge-$\epsilon$ $-$ Cloud+MTP & \begin{tabular}[t]{@{}r@{}}$+0.5$\\$[-0.7, +1.8]$\end{tabular} & \begin{tabular}[t]{@{}r@{}}$-0.4$\\$[-2.1, +1.5]$\end{tabular} & \begin{tabular}[t]{@{}r@{}}$+2.8$\\$[+0.1, +6.2]$\end{tabular} & \begin{tabular}[t]{@{}r@{}}$-0.3$\\$[-0.8, +0.2]$\end{tabular} & \begin{tabular}[t]{@{}r@{}}$-0.2$\\$[-2.5, +2.1]$\end{tabular} & \begin{tabular}[t]{@{}r@{}}$+0.3$\\$[-4.5, +5.2]$\end{tabular} \\[2pt]
\bottomrule
\end{tabular*}

  \caption{Repeated SB quality contrasts, in judge points: mean difference and paired 95\% CI. All: non-QA first turns, $n=398$ for \sysname--Cloud+MTP and $n=396$ otherwise. Task sizes are 80 except summarization (78) and math in comparisons involving SpecEdge-$\epsilon$ (78). These 15 task-wise contrasts are exploratory and not multiplicity-adjusted. A CI spanning zero does not establish equivalence.}
  \label{tab:quality-ci}
\end{table*}

\paragraph{Bootstrap procedure.}
The release uses 2,000 percentile-bootstrap draws with seed 20261011 restarted per contrast, with question IDs sorted as strings.
Both MT-bench turns are resampled together for efficiency.
Repeated quality averages all three available scores per question before complete-case pairing and resampling.
These intervals reflect variation across recorded questions conditional on the observed runs; they are not an estimate of variability across arbitrary deployments.
The separate run SD combines generation and judging variability because each output has only one judge evaluation.
Intervals may differ slightly from intermediate logs whose random-number streams were consumed in another order.

\paragraph{QA diagnostics.}
The question-only judge is not a reference-based QA accuracy test, so we report QA separately from the non-QA aggregate.
Our NQ-Open reference matching procedure matched 43 of the 80 QA questions; the remaining 37 were unmatched.
Using the same normalized reference-containment rule as for RAG, Cloud+MTP scores 30.2\%, \sysname\ 30.2\%, \sysname\ exact 25.6\%, and SpecEdge-$\epsilon$ 29.5\% on these 43 items.
For the two repeated systems, each question's three binary scores are averaged; the other systems use one answer.
The \sysname--Cloud+MTP difference is 0.0 percentage points (95\% CI: $-13.2$ to $+10.9$).
Unmatched questions are omitted from this diagnostic rather than counted as errors; containment is not strict exact match.
The question-only QA judge means, covering all 80 questions, are 49.1 for \sysname\ and 45.1 for Cloud+MTP; they do not establish reference-answer correctness.

\paragraph{Reference metrics.}
The optional reference-metric script computes WMT14 corpus BLEU, CNN/DailyMail ROUGE-L, GSM8K numeric-answer accuracy, RAG contains-gold accuracy, and InfiniteBench ROUGE-L-Sum.
The RAG measure checks normalized reference containment, not full answer correctness.
For a descriptive check on \sysname's first generation, GSM8K accuracy is 95.0\% for both \sysname and Cloud+MTP; RAG accuracy is 65.7\% versus 67.1\%; ROUGE-L is 19.2 versus 19.8; BLEU is 33.9 versus 33.7.
These values refer to one output set, not the three-run judge aggregate.
On 32K summaries, the recorded ROUGE-L-Sum differences are $-1.0$ [$-2.1$, $+0.1$] for \sysname and $-0.8$ [$-1.8$, $+0.1$] for SpecEdge-$\epsilon$.
For SpecEdge-style and PD, the differences are $-0.2$ [$-1.0$, $+0.4$] and $-0.8$ [$-2.0$, $+0.2$], on 50 and 49 paired summaries, respectively.

\begin{table*}[t]
  \centering\small\setlength{\tabcolsep}{4pt}
  \begin{tabular*}{\textwidth}{@{\extracolsep{\fill}}lrrrr@{}}
\toprule
Configuration & F1 & $\Delta$ F1 & $\Delta$ recall & $\Delta$ precision \\
\midrule
Cloud+MTP & 29.6 & -- & -- & -- \\
SpecEdge-style & 31.3 & $+1.6$ $[-1.2, +4.4]$ & $+1.3$ $[-0.7, +3.3]$ & $+5.8$ $[+1.3, +10.2]$ \\
SpecEdge-style + PD & 29.1 & $-0.9$ $[-4.1, +2.1]$ & $+1.1$ $[-1.0, +3.0]$ & $-1.7$ $[-7.1, +4.1]$ \\
SpecEdge-$\epsilon$ & 28.5 & $-1.2$ $[-3.6, +1.4]$ & $-2.2$ $[-4.3, 0.0]$ & $+6.2$ $[+2.0, +10.6]$ \\
\sysname & 30.0 & $+0.4$ $[-2.3, +3.1]$ & $-0.1$ $[-2.3, +2.1]$ & $+1.9$ $[-2.6, +6.2]$ \\
\sysnametwo & 28.0 & $-1.7$ $[-4.3, +1.0]$ & $-1.9$ $[-4.1, +0.3]$ & $+4.3$ $[-0.3, +8.9]$ \\
\sysname\ exact & 29.7 & $0.0$ $[-2.5, +2.6]$ & $+0.1$ $[-1.8, +2.1]$ & $+1.8$ $[-3.5, +6.9]$ \\
\sysname\ server head & 29.0 & $-0.6$ $[-3.4, +2.2]$ & $+0.4$ $[-2.1, +3.0]$ & $+0.7$ $[-3.2, +4.6]$ \\
\bottomrule
\end{tabular*}

  \caption{HELMET model-based evaluation on 50 \infsum\ 32K summaries per configuration (49 for PD), one generated answer and one metric evaluation per answer. Values are percentage points; differences are against Cloud+MTP with paired 95\% CIs. F1 includes the fluency factor. PD differences use the 49 common summaries, so they need not equal differences of displayed marginal means. Recall measures key-point coverage; precision measures support by the reference expert summary. These are not human factuality judgments against full novels.}
  \label{tab:long-quality}
\end{table*}

The HELMET evaluation retains its book-summary prompts and key points, using GPT-5.4 as the judge.
One metric evaluation contains separate fluency, recall, and precision requests, not three independent repeated evaluations of the same metric.
The script requests temperature 0.1, top-$p$ 0.9 and seed 42, omitting a field only if the API rejects it.
The reported comparisons use the recorded scores, not new API calls.

\section{Failure Cases}
\label{app:failures}

\paragraph{Server cost.}
Among 480 questions, \sysname has 2, 1, and 0 cases with $\Delta g\ge0$ in its three runs.
The non-saving cases are short RAG answers of approximately 40 tokens.
This observation illustrates an unfavorable operating condition; it does not establish that output length alone causes every cost difference.

\paragraph{Quality-screened cases.}
We apply a fixed $\pm20$-point screen to the difference between each question's three-run \sysname mean and Cloud+MTP's single score.
This selects 13 decreases and 12 increases.
The following categories are our own judgments, not an independent adjudication or the overall error rate.

\begin{table}[t]
  \centering\small\setlength{\tabcolsep}{4pt}
  \begin{tabular*}{\columnwidth}{@{\extracolsep{\fill}}lrr@{}}
    \toprule
    Our classification & Decreases & Increases \\
    \midrule
    Substantive \sysname error & 6 & -- \\
    Minor difference & 2 & 1 \\
    Score/content discrepancy & 5 & 9 \\
    Cloud+MTP defect & -- & 2 \\
    \midrule
    Selected cases & 13 & 12 \\
    \bottomrule
  \end{tabular*}
  \caption{Manual inspection of large score changes. No scored item is removed from the quantitative analysis.}
  \label{tab:failures}
\end{table}

Seven selected decreases have one especially low-scoring \sysname generation.
In one inspected question (559), identical answer text received scores of 15 and 95, showing that at least some large changes can originate in the evaluator rather than different text.
This does not establish that all score discrepancies are judge errors.

\section{System Metrics and Reproducibility}
\label{app:sysmetrics}

\paragraph{Metric definitions.}

\begin{table*}[t]
  \centering\small\setlength{\tabcolsep}{4pt}
  \begin{tabular*}{\textwidth}{@{\extracolsep{\fill}}lrrrrr@{}}
\toprule
Configuration & $g$ (ms/token) & Tok/s & Draft ms/token & Tokens/GPU-busy s & GPU-h/1M tokens \\
\midrule
\multicolumn{6}{@{}l}{\emph{SB-480}} \\
Cloud+MTP & 7.81 & -- & -- & 128.1 & 2.17 \\
SpecEdge-style & 6.20 & 16.1 & 42.8 & 161.3 & 1.72 \\
SpecEdge-style + PD & 5.94 & 12.1 & 40.1 & 168.4 & 1.65 \\
SpecEdge-$\epsilon$ & 5.50 & 18.9 & 37.0 & 181.9 & 1.53 \\
\sysname & 4.80 & 35.0 & 12.1 & 208.5 & 1.33 \\
\sysnametwo & 4.32 & 29.0 & 18.2 & 231.5 & 1.20 \\
\sysname\ exact & 5.64 & 28.2 & 14.7 & 177.2 & 1.57 \\
\sysname\ server head & 6.67 & 36.1 & 9.2 & 149.9 & 1.85 \\
On-device 8K & -- & 26.4 & 36.8 & -- & -- \\
\midrule
\multicolumn{6}{@{}l}{\emph{\infsum\ 32K}} \\
Cloud+MTP & 23.91 & -- & -- & 41.8 & 6.64 \\
SpecEdge-style & 23.20 & 10.0 & 60.3 & 43.1 & 6.45 \\
SpecEdge-style + PD & 23.18 & 8.2 & 58.9 & 43.1 & 6.44 \\
SpecEdge-$\epsilon$ & 20.16 & 12.4 & 48.9 & 49.6 & 5.60 \\
\sysname & 16.92 & 21.0 & 16.4 & 59.1 & 4.70 \\
\sysnametwo & 16.27 & 18.8 & 24.9 & 61.5 & 4.52 \\
\sysname\ exact & 21.07 & 16.8 & 21.1 & 47.5 & 5.85 \\
\sysname\ server head & 23.00 & 19.8 & 12.8 & 43.5 & 6.39 \\
\bottomrule
\end{tabular*}

  \caption{Metrics recomputed from the finalized per-request records. Repeated SB configurations average each run's metric; other entries use one run. Tokens per GPU-busy second equal $1000/g$ per run, and GPU-hours per million tokens equal $g/3.6$ when $g$ is in milliseconds. These are execution-efficiency measures for one client, not multi-client serving throughput or priced service costs. --: no applicable server or device measurement.}
  \label{tab:sysmetrics}
\end{table*}

Table~\ref{tab:sysmetrics} separates GPU execution efficiency from user-visible delivery.
Per request, $g$ includes \texttt{server\_ms} plus \texttt{prefill\_ms}.
Cloud+MTP already includes prefill in the former, so the latter is zero.
Mobile draft time is \texttt{phone\_ms} minus \texttt{verify\_wait\_ms} and \texttt{phone\_prefill\_ms}, divided by output tokens.
All ratios pool numerators and denominators before averaging across repeated runs.
For PD, subtracting the verification wait also excludes any drafting overlapped with that wait, so the mobile-time column is not its total active compute.

\paragraph{Missing records and latency traces.}
\textbf{PD recording coverage.}
The SB PD run contains 558 of 560 scheduled turns; its paired efficiency contrast uses 479 common questions.
Two recorded turns (questions 559 and 560) have no assigned PRIME prefill and retain \texttt{prefill\_ms}=0.
We do not impute these times, so reported PD GPU time may be slightly underestimated.
On 32K, PD completes 49 of 50 requests; its efficiency and quality contrasts use those 49 common questions.
Other finalized long-summary configurations have 50 records.

\paragraph{First-token latency.}
The recorded single-run latency export reports \sysname mean/p90 TTFT of 846/1,922\,ms on SB and 4,604/4,842\,ms on 32K summaries.
For \sysname-2P these are 967/2,147\,ms and 3,759/3,908\,ms, and for \sysname exact 920/2,050\,ms and 4,014/4,344\,ms.
Recorded SB summaries put network and serialization at roughly 40--58\,ms per round, after subtracting server decoding from the verification wait; this is not a standalone link-RTT measurement. These are individual-run traces, not three-run averages; other finalized configurations do not have a corresponding released per-event latency trace.
The initial head-state transfer can contain up to 1,024 positions.
Consequently, first-token latency must be considered separately from average delivery speed.

\section{Additional Cross-Family Measurements}
\label{app:crossfamily}

The SLM component of the \hybriddrafter is not tied to Gemma~4.
In additional runs, our \slmdrafter drafted without retraining on a Galaxy Z Fold8 smartphone, which shares the S26 Ultra's SM8850 chipset, for Muse-Glimmer-30B in NVFP4~\citep{meta2026museglimmer,nvidia2026museglimmernvfp4}, a verifier of another family whose tokenizer it is mapped to through a token bridge.
Glimmer's own DFlash2 head~\citep{chen2026dflash,incoai2026museglimmerdflash2}, which reads five of its layers and was not deployed on the device, drafted eight-token chains on the server, as in our \serverhead variant \unskip{}.
Over all 480 SB questions, a SpecEdge-style tree of four SLM passes and 127 nodes (128 \verifier rows, against 33 in Table~\ref{tab:configs}) reached an MAT of 3.16, and a 32-node tree of the chain and a single SLM pass reached 4.07.

These runs used another verifier, server GPU, and engine, and answers of up to 2,048 tokens, so they are not comparable with Table~\ref{tab:main}, but they show that the \slmdrafter carries over to another model family through a token bridge.
A head that the mobile device can run for such families would move the whole \hybriddrafter, not only its SLM half, to the device, which we leave to future work.

\section{Artifacts, Licenses and Compute}
\label{app:artifacts}

\begin{table*}[t]
  \centering
  \small
  \setlength{\tabcolsep}{5pt}
  \begin{tabular}{@{}>{\raggedright\arraybackslash}p{9.4cm}ll@{}}
    \toprule
    Artifact & Use & License \\
    \midrule
    \multicolumn{3}{@{}l}{\emph{Models}} \\
    Gemma~4 26B-A4B-it~\citep{gemmateam2026gemma4}, 25.2B parameters (3.8B active), AWQ 4-bit~\citep{lin2024awq}: \texttt{cyankiwi/gemma-4-26B-A4B-it-AWQ-4bit} & \verifier & Apache 2.0 \\
    Its MTP head, 420M parameters: \texttt{google/gemma-4-26B-A4B-it-assistant} & \mtpdrafter & Apache 2.0 \\
    Gemma~4 E2B-it, Google's QAT checkpoint, 2.3B effective parameters: \texttt{google/gemma-4-E2B-it-qat-mobile-transformers}~\citep{google2026gemma4e2bqat} & \slmdrafter & Apache 2.0 \\
    Muse-Glimmer-30B~\citep{meta2026museglimmer}, 29.6B parameters, NVFP4: \texttt{nvidia/Muse-Glimmer-30B-NVFP4}~\citep{nvidia2026museglimmernvfp4} & \begin{tabular}[t]{@{}l@{}}cross-family \verifier\\(App.~\ref{app:crossfamily})\end{tabular} & Apache 2.0 \\
    Its DFlash2 head~\citep{chen2026dflash}, 2.8B parameters: \texttt{incoai/Muse-Glimmer-30B-DFlash2}~\citep{incoai2026museglimmerdflash2} & its server-side head & Apache 2.0 \\
    GPT-5.4~\citep{openai2026gpt54} (snapshot 2026-03-05), through the Azure OpenAI service & judge & API terms of use \\
    \midrule
    \multicolumn{3}{@{}l}{\emph{Data}} \\
    SB~\citep{xia2024unlocking} & test & Apache 2.0 \\
    \quad with MT-bench~\citep{zheng2023judging} & & Apache 2.0 \\
    \quad with WMT14 German--English~\citep{bojar2014findings} & & none stated \\
    \quad with CNN/DailyMail~\citep{nallapati2016abstractive} & & Apache 2.0 \\
    \quad with Natural Questions~\citep{kwiatkowski2019natural} & & CC BY-SA 3.0 \\
    \quad with DPR passages~\citep{karpukhin2020dense} & & CC BY-NC 4.0 \\
    \quad with GSM8K~\citep{cobbe2021training} & & MIT \\
    InfiniteBench~\citep{zhang2024infinitebench}, its English summarization task (En.Sum, \infsum) & test & MIT \\
    UC~\citep{ding2023enhancing,tunstall2024zephyr}: UC-val from its \texttt{test\_sft} split, UC-tr its \texttt{train\_sft} split & validation, keep set, QAT$^b$ & MIT \\
    \midrule
    \multicolumn{3}{@{}l}{\emph{Software}} \\
    PyTorch~\citep{paszke2019pytorch} & server, training & BSD-3-Clause \\
    Transformers~\citep{wolf2020transformers} & model code & Apache 2.0 \\
    SpecEdge server~\citep{park2025specedge}, the base of our \verifier engine & \verifier & research use$^a$ \\
    NNTrainer~\citep{moon2024frontier} with its CausalLM application, the base of our mobile runtime & \slmdrafter, \mtpdrafter & Apache 2.0 \\
    QAIRT 2.47.1~\citep{qualcomm2026qairt} & NPU graphs & AI Stack License \\
    vLLM~\citep{kwon2023efficient} with its Marlin kernels~\citep{frantar2025marlin} & \verifier kernels & Apache 2.0 \\
    FlashInfer~\citep{ye2025flashinfer} & \verifier kernels & Apache 2.0 \\
    HELMET~\citep{yen2025helmet}, preprocessing and scoring code & \infsum evaluation & MIT \\
    \bottomrule
  \end{tabular}
  \caption{Artifacts we used, with the licenses that their distributors declare on Hugging Face, on GitHub or in the license file. Model checkpoints are Hugging Face repositories. $^a$Research, evaluation and education; commercial deployment needs a separate license. $^b$UC-tr: keep-set counts and 60 disjoint prompts for final head-QAT and activation calibration (Appendix~\ref{app:qat}).}
  \label{tab:artifacts}
\end{table*}

Table~\ref{tab:artifacts} shows the models, data and software we used, with the license that each distributor declares \unskip{}.
We use each artifact as intended: the models for inference, the benchmarks for evaluation, and UC only for the uses in Table~\ref{tab:artifacts}.
The one model we train, the \mtpdrafter, is a derivative that its Apache~2.0 license permits.
The SpecEdge engine permits research and evaluation but needs a separate license for commercial deployment, and the Dense Passage Retrieval (DPR) passages in SB are licensed for noncommercial use only.
Our use is noncommercial research.
The test data are public and in English, except for the German source sentences of SB's translation category.

For the main Gemma deployment, weight training is the MTP head's QAT: 100 steps of about 1.5\,s each on one NVIDIA DGX Spark.
The final-run records account for 5.04 GPU-busy hours and 31.00 cumulative mobile-request hours.
These sums are not the full development budget: they exclude setup, idle time, failed requests absent from the records, calibration, validation sweeps, and additional cross-family experiments.
The final GPU is a 96\,GB RTX PRO 6000 in an AWS g7e.2xlarge instance.

\paragraph{AI assistance.}
We acknowledge the use of an AI assistant for language editing of the manuscript and for help with writing the experiment and analysis scripts.

\fi

\end{document}